\documentclass[11pt]{article}

\PassOptionsToPackage{x11names}{xcolor}

\usepackage[preprint]{acl}

\definecolor{Gray}{HTML}{BEBEBE}

\definecolor{citeblue}{HTML}{0B5394}
\definecolor{refmaroon}{HTML}{8B0000}
\definecolor{resourceskyblue}{HTML}{1E88E5}
\hypersetup{
  colorlinks=true,
  citecolor=citeblue,
  linkcolor=refmaroon,
  urlcolor=citeblue,
}

\usepackage{times}
\usepackage{latexsym}

\usepackage[T1]{fontenc}

\usepackage[utf8]{inputenc}

\usepackage{microtype}

\usepackage{inconsolata}

\usepackage{graphicx}

\usepackage{url}
\usepackage{booktabs}
\usepackage{amsmath}
\usepackage{amsfonts}
\usepackage{nicefrac}

\usepackage{multirow}
\usepackage{colortbl}
\usepackage{array}
\usepackage{caption}

\usepackage[most]{tcolorbox}
\definecolor{lightbeige}{HTML}{F5F5DC}

\newcolumntype{C}[1]{>{\centering\arraybackslash}p{#1}}

\usepackage{amssymb}

\usepackage{etoc}

\title{Can MLLMs Read the Room? A Multimodal Benchmark for Assessing Deception in Multi-Party Social Interactions}

\author{
  \textbf{Caixin Kang},
  \textbf{Yifei Huang}\thanks{Corresponding author.},
  \textbf{Liangyang Ouyang} \\
  \textbf{Mingfang Zhang},
  \textbf{Ruicong Liu},
  \textbf{Yoichi Sato} \\
  The University of Tokyo, Tokyo, Japan \\
  \small{\texttt{\{cxkang,hyf,ysato\}@iis.u-tokyo.ac.jp}}
}

\begin{document}

\hypersetup{linkcolor=black}
\maketitle
\hypersetup{linkcolor=refmaroon}

\etocdepthtag.toc{mtmain}

\begin{abstract}

Despite their advanced reasoning capabilities, state-of-the-art Multimodal Large Language Models (MLLMs) demonstrably lack a core component of human intelligence: the ability to `read the room' and assess deception in complex social interactions. 
To rigorously quantify this failure, we introduce a new task, \textbf{Multimodal Interactive Deception Assessment (MIDA)}, and present a novel multimodal dataset providing synchronized video and text with verifiable ground-truth labels for every statement.
We establish a comprehensive benchmark evaluating 12 state-of-the-art open- and closed-source MLLMs, revealing a significant performance gap: even powerful models like GPT-4o struggle to distinguish truth from falsehood reliably. Our analysis of failure modes indicates that these models fail to effectively ground language in multimodal social cues 
and lack the ability to model what others know, believe, or intend,
highlighting the urgent need for novel approaches to building more perceptive and trustworthy AI systems.
To take a step forward, we design a Social Chain-of-Thought (SoCoT) reasoning pipeline and a Dynamic Social Epistemic Memory (DSEM) module. Code and data will be released through our \href{https://github.com/kkkcx/Can-MLLMs-Read-the-Room}{\textcolor{resourceskyblue}{project repository}} and \href{https://huggingface.co/datasets/Ukpkmkkk/MIDA}{\textcolor{resourceskyblue}{dataset page}}.

\end{abstract}
\begin{figure*}[t]
  \centering
   \includegraphics[width=1.00\linewidth]{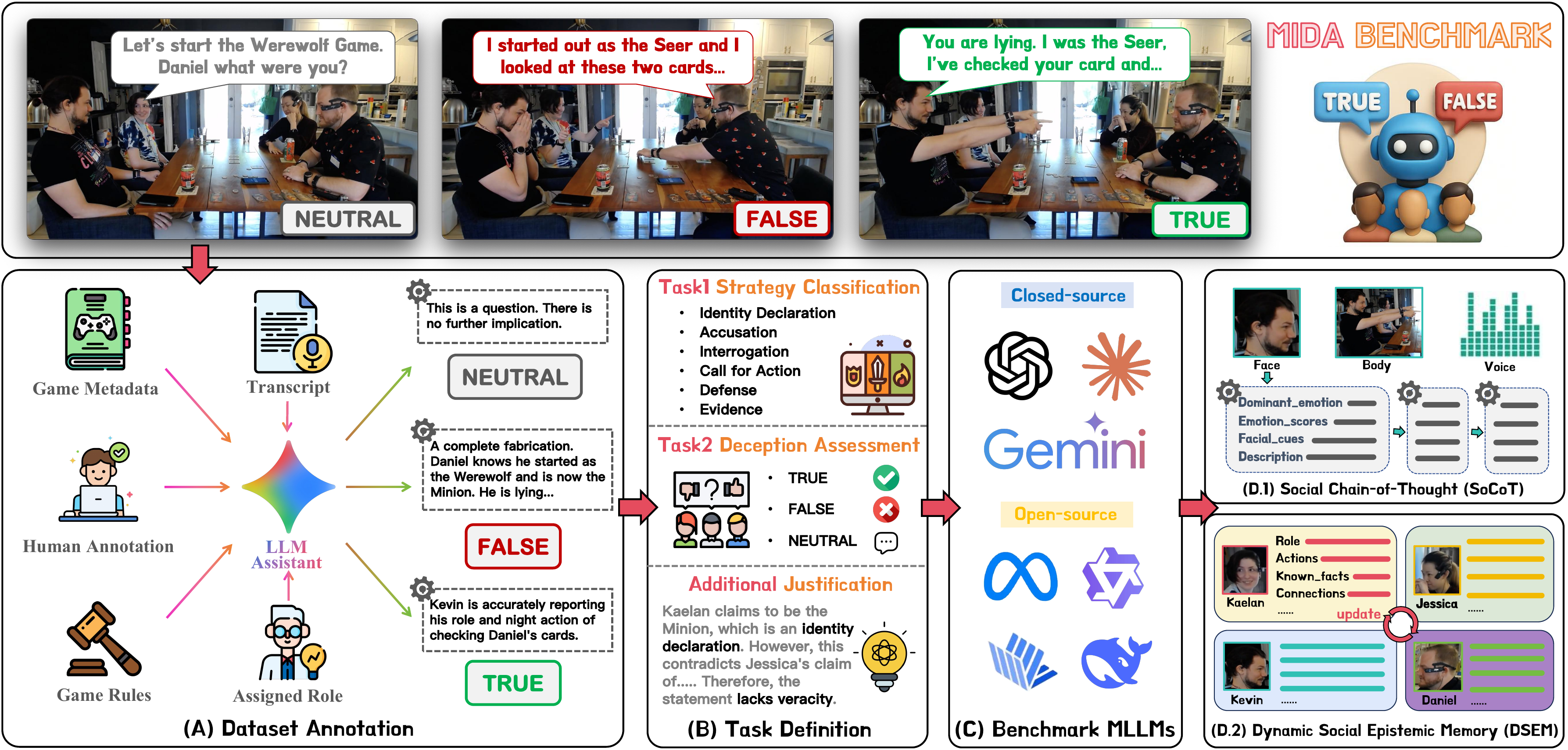}
   \caption{
   \textbf{Overview of the MIDA Benchmark and New Modules for Deception Reasoning. }
   (A) A semi-automated process where game transcripts, rules, and metadata are used by an LLM assistant to generate veracity labels. (B) The MIDA task. It requires models to perform reasoning from Persuasive Strategy Classification to Deception Assessment with a supporting justification. (C) The evaluated MLLMs. (D) Proposed social-reasoning modules: (D.1) SoCoT grounds inference in multimodal cues, while (D.2) DSEM tracks participants' social states.
   }
    \label{fig:pipeline}
     \vspace{-3ex}
\end{figure*}

\section{Introduction}
\label{sec:intro}

Deception, a pervasive social phenomenon~\cite{ekman2009telling}, profoundly impacts interpersonal relationships, economic activities, and even public safety. 
Accurately identifying deceptive intent in discourse is crucial for understanding the complexities of human interactions. It also presents a core challenge for building more intelligent and secure AI systems, such as advanced conversational agents and content moderation platforms \cite{picard1997affective}. To foster a future where AI and humans can collaborate seamlessly, equipping AI with sophisticated social perception and reasoning capabilities is indispensable.

While deception detection has garnered research attention, existing work largely suffers from three major limitations. 
\textbf{(i) Lack of interactional context.} Prior studies often operate in isolation, analyzing single text snippets~\cite{ott2011finding}, unidirectional speech videos~\cite{perez2015deception, van2015deception}, or independent physiological signals~\cite{langleben2002brain, vrij2010pitfalls}. However, real-world deception is not a static act but a dynamic, interactive process. 
\textbf{(ii) Simplification of social complexity. }
While prior work has made progress in two-person settings~\cite{soldner2019box}, their interaction patterns are relatively structured (one person describes, another guesses). 
This fails to capture the complexity of authentic social deception, which frequently unfolds in messy, multi-party networks of shifting alliances, rivalries, and group pressure common to human social dynamics~\cite{meta2022human}.
\textbf{(iii) Scarcity of verifiable ground truth.} A pivotal obstacle is the lack of publicly available datasets with verifiable ground truth. In real-world scenarios, objectively annotating the precise moments of deception is often impossible~\cite{perez2015deception}. 

To address these limitations, we introduce the social deduction game \textit{Werewolf}, 
where players must conceal their secret roles, using deceptive communication to mislead others. By day, everyone votes to eliminate a player, while at night, Werewolves and other special roles secretly perform their night actions to achieve their hidden goals.
This complex multi-party game serves as a controlled yet ecologically valid environment that elicits natural, high-stakes deception. Crucially, its rule-based nature provides objective, deterministic ground truth for every statement made~\cite{lai2023werewolf}.
Building on this paradigm, we formalize a new task Multimodal Interactive Deception Assessment (MIDA). We construct a novel multimodal benchmark dataset totaling 2,360 utterances. Our semi-automated pipeline involves manually annotating crucial \textbf{``night actions"} to mitigate the uncertainty, then using an LLM assistant to parse events, and rigorously verifying initial veracity labels against the game's ground truth, ensuring high-fidelity, verifiable annotations.

We conduct a comprehensive benchmark of 12 state-of-the-art open- and closed-source MLLMs on MIDA, revealing significant performance deficiencies. Our analysis categorizes the following core failure modes:
1) Difficulty in distinguishing salient social signals from distracting noise; 
and 2) A lack of functional "Theory of Mind" \cite{kosinski2023theory} — the ability to internally model what others know, believe, or intend, hindering the inference of hidden beliefs and strategic intent.
In essence, current MLLMs function as powerful knowledge engines but fall short as competent social agents.

In response to the failure modes, we propose two modules for deception reasoning: a Social Chain-of-Thought (SoCoT) pipeline, which provides multimodal interpretable step-wise inference, forcing explicit grounding in visual and acoustic cues. And a Dynamic Social Epistemic Memory (DSEM) module, which offers persistent and structured cognitive context to maintain and update each participant's evolving beliefs and strategic intentions. Experimental results confirm that these modules improve MLLM performance on the MIDA benchmark, moving them toward a more faithful modeling of human-like social intelligence. In summary, our contributions are as follows:

\begin{itemize}
\vspace{-1ex}
\item \textbf{New Task \& Dataset:} We introduce the \textbf{MIDA} (Multimodal Interactive Deception Assessment) task and a novel, large-scale dataset with verifiable ground truth.

\vspace{-1ex}
\item \textbf{Comprehensive Benchmark:} We benchmark 12 leading MLLMs on MIDA, revealing a huge performance gap and analyzing key failure modes, including a lack of Theory of Mind and poor multimodal social signals grounding.

\vspace{-1ex}
\item \textbf{Novel Modules:} We propose the Social Chain-of-Thought (SoCoT) reasoning pipeline and the Dynamic Social Epistemic Memory (DSEM) module to improve MLLMs' performance on social deception reasoning, advocating for more cognitively grounded AI systems.

\end{itemize}

\section{Related Work}
\label{sec:related_work}
\vspace{-1ex}
\subsection{Multimodal Social Interaction}

Multimodal social interaction integrates language, acoustic, and visual signals to model nuanced human behavior. Foundational corpora such as AMI~\cite{carletta2005ami,pentland2010honest}, CMU-MOSI~\cite{zadeh2016mosi}, and CMU-MOSEI~\cite{zadeh2018multi} target observable group dynamics and affective states (e.g., sentiment, engagement) from monologues or meetings. More recent Werewolf-grounded studies analyze persuasion~\cite{lai2023werewolf}, social interactions~\cite{lee2024modeling}, gestures~\cite{cao2025socialgesture}, and online interaction~\cite{li2025towards}, establishing strong baselines for multimodal fusion. Our work diverges by targeting a latent, intentional behavior, \textit{deception}, which requires reasoning beyond surface affect to a speaker's hidden knowledge, intent, and strategic motives in a high-stakes adversarial context.
\vspace{-1ex}

\subsection{Deception Detection}

Early deception detection focused on single modalities: text~\cite{ott2011finding,li2014towards,fornaciari2013automatic}, vocal cues~\cite{montacie2016prosodic,litvinova2017building}, facial micro-expressions and gaze~\cite{ekman2009telling,yan2014casme,perez2015verbal}, and physiological signals~\cite{langleben2002brain}. Multimodal approaches followed, with the Real-Life Trial corpus~\cite{perez2015deception} contributing monologue-style videos and ``Box of Lies''~\cite{zadeh2019boxoflies} introducing a two-person conversational setting. MIDA pushes this frontier in two directions: (i) from non-interactive monologues and structured dyadic dialogues to messy, multi-party interactions with shifting alliances and rivalries; and (ii) from ambiguous human-judged labels to objective, rule-derived GT for every utterance, addressing a long-standing annotation challenge.
\vspace{-1ex}

\subsection{Computational Modeling of Deduction Games}

Complex games have long served as testbeds for AI, from perfect-information settings (Chess, Go~\cite{silver2016mastering}, Poker~\cite{brown2018superhuman}) to social-reasoning games like Diplomacy~\cite{meta2022human} and text-based Werewolf~\cite{toriumi2016ai}. Recent LLM-based studies in the Werewolf domain focus on agent design and strategic play~\cite{xu2023exploring,wu2024enhance} or arena win-rate evaluation~\cite{bailis2024werewolf}. We take a fundamentally different angle: rather than building a game-playing agent, we use Werewolf as a multimodal benchmark to ask \textit{whether MLLMs can read the room by grounding language in visual and auditory cues}. To our knowledge, MIDA is the first benchmark to bridge social-deduction games and multimodal deception assessment in complex, multi-party settings.

\begin{figure*}[t]
  \centering
   \includegraphics[width=0.80\linewidth]{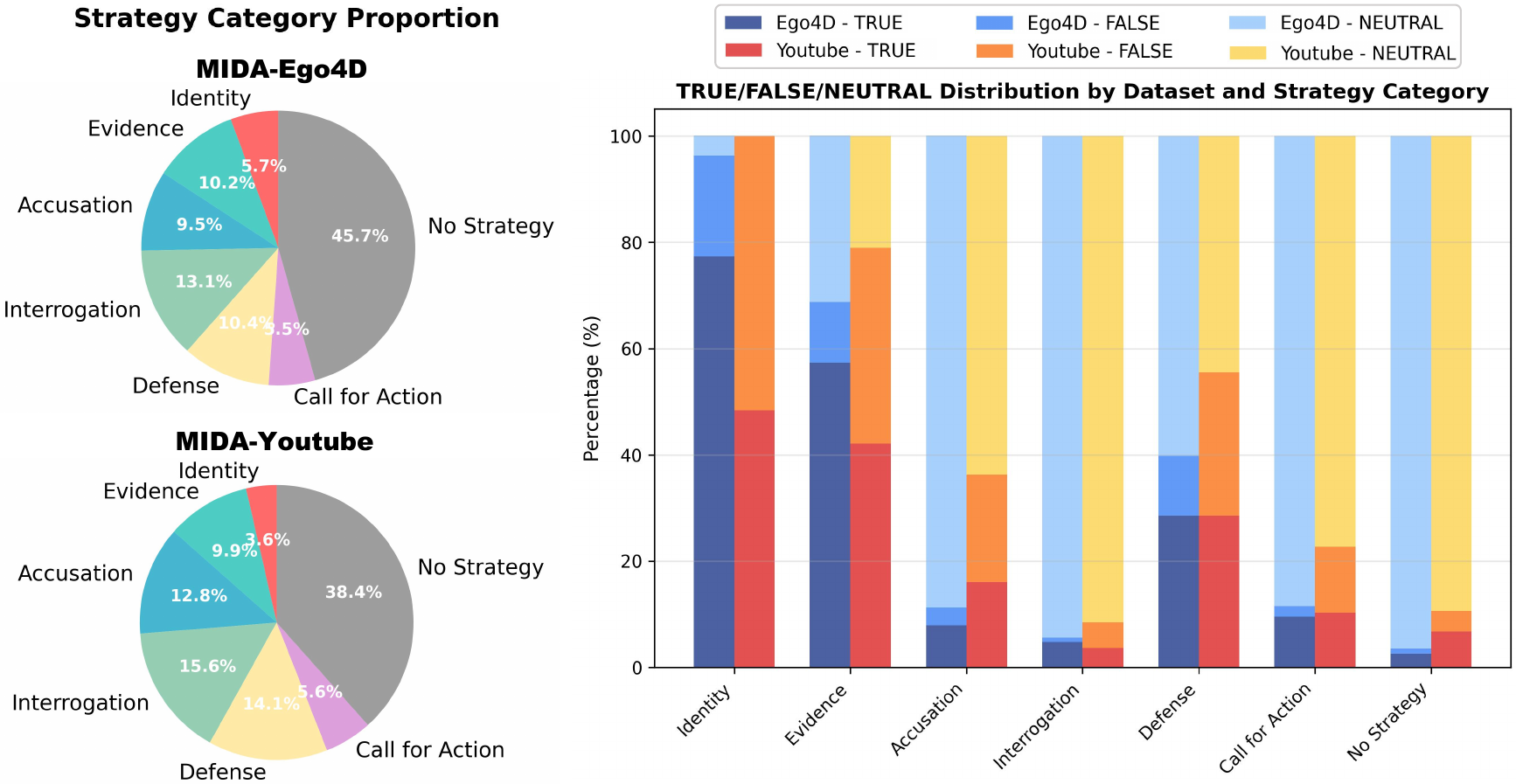}
    \vspace{-1ex}
   \caption{Persuasive Strategy proportion and distribution of veracity labels across Persuasive Strategy categories in MIDA-Ego4D/MIDA-YouTube.}
    \label{fig:proportion_chart}
     \vspace{-3ex}
\end{figure*}

\vspace{-1ex}
\section{The MIDA Benchmark}
\label{sec:dataset}

\vspace{-1ex}
\subsection{Dataset Curation and Annotation}

\noindent\textbf{Data Sources.}
Our dataset is built upon two corpora featuring the game ``One Night Ultimate Werewolf'', totaling 2,360 utterances:
\begin{itemize}
    \item \textbf{MIDA-Ego4D}: We use a subset of the Ego4D social dataset~\cite{grauman2022ego4d}, consisting of 40 game sessions recorded from a third-person perspective to ensure all participants are visible. Our annotated test set comprises 819 utterances from a split defined in prior work~\cite{lai2023werewolf}.
    \item \textbf{MIDA-YouTube}: This corpus is curated from 151 game videos collected from YouTube~\cite{lai2023werewolf}, from which a subset of 1541 utterances was selected for our detailed veracity annotation.
\end{itemize}

\noindent\textbf{Annotation Pipeline.}
As shown in Figure~\ref{fig:pipeline} (A), to address the inherent ambiguity of deducing night events from start and end roles alone, we introduce the manual annotation of \textbf{``night actions''}, which records the precise target of every action taken during the night phase (e.g., which player the Robber targeted).
We then introduce a novel semi-automated pipeline, employing Gemini-2.5-Pro assigned the role of ``Expert Analyst", to analyze game transcripts against the game rules and deterministic ground truth from the annotated night actions and game metadata (e.g., roles, voting results) and generate objective labels.

\noindent\textbf{Annotation Principles.}
The cornerstone of our annotation is that \textit{deception is judged based on the speaker's private knowledge state at the moment of utterance}. To operationalize this, we develop a detailed \textbf{rulebook} (see Appendix~\ref{appendix:annotation_eval}).
Formally, let $K_s^t$ denote speaker $s$'s private knowledge state at time $t$, induced by their role, night actions, and observations, and let $\varphi$ denote an utterance's propositional content. We define
\[
\operatorname{label}(\varphi;K_s^t)=
\begin{cases}
\texttt{TRUE},    & K_s^t \models \varphi,\\
\texttt{FALSE},   & K_s^t \models \neg\varphi,\\
\texttt{NEUTRAL}, & \text{otherwise}.
\end{cases}
\]
The \texttt{NEUTRAL} case covers guesses, opinions, questions, hypotheticals, and other statements that are not falsifiable from $K_s^t$; Appendix~\ref{appendix:annotation_details} specifies the recurring boundary cases.
To ensure data quality, we conduct a human validation study on a subset (20\%) of the MIDA-Ego4D data, achieving a high human-LLM agreement accuracy of 96.6\% (see Appendix~\ref{appendix:human_validation} for details).

\subsection{Task Definition and Evaluation}
We formally define the MIDA task as follows: For a player's utterance, we provide the corresponding image and audio, together with the conversational history and the game rules. The objective is to predict its veracity label from the set $\{\texttt{TRUE, FALSE, NEUTRAL}\}$.
At test time, semantic game-state metadata, including player roles, night actions, and voting outcomes, is never provided to the model. Speaker bounding boxes are used only in offline preprocessing to crop face and body regions. Neither the coordinates nor any game-state semantics are included in the model prompt.
For evaluation, 
the MLLM is first prompted to identify the presence of six persuasive strategies (\textit{Identity Declaration}, \textit{Accusation}, \textit{Defense}, \textit{Evidence}, \textit{Interrogation}, \textit{Call for Action}) defined in prior work~\cite{lai2023werewolf}. Subsequently, it must determine the final veracity label, together with additional justification.
All the output is required in a structured JSON format. 
The detailed prompt is provided in Appendix~\ref{appendix:annotation_eval}.

\subsection{Dataset Analysis}
\label{sec:dataset_analysis}

\noindent\textbf{Data Imbalance and Category Bias.}
As Figure~\ref{fig:proportion_chart} shows, the multi-party discourse is dominated by \texttt{NEUTRAL} utterances and factual claims are sparse overall. Over 90\% of statements in \textit{Interrogation} and \textit{No Strategy} carry no verifiable claim, whereas \texttt{TRUE}/\texttt{FALSE} statements concentrate in high-stakes categories such as \textit{Identity Declaration} and \textit{Evidence}. A model must therefore first separate the fact-checkable signal from the surrounding neutral conversation before deception can be assessed. 

\noindent\textbf{Dataset Variation and Strategic Divergence.}
The two corpora form a natural difficulty gradient: MIDA-Ego4D collects novice players, while MIDA-YouTube features experienced ones, and the gap is reflected in both strategy and deception. Experts speak more strategically, producing fewer non-strategic utterances than novices (38.4\% vs.\ 52.5\%), and engage in markedly more sophisticated deception, false \textit{Evidence} claims occur nearly $3\times$ more often (36.8\% vs.\ 11.5\%), and identity lies more than twice as frequently (51.6\% vs.\ 18.9\%).

Although grounded in Werewolf, MIDA's design principles and reasoning modules may extend to other social-reasoning settings. Semantic game-state metadata is used only for ground-truth labeling and is withheld at inference; speaker bounding boxes are used solely for offline visual cropping. A full generalizability discussion is in Appendix~\ref{sec:generalizability}.
\begin{figure*}[t]
  \centering
   \includegraphics[width=0.99\linewidth]{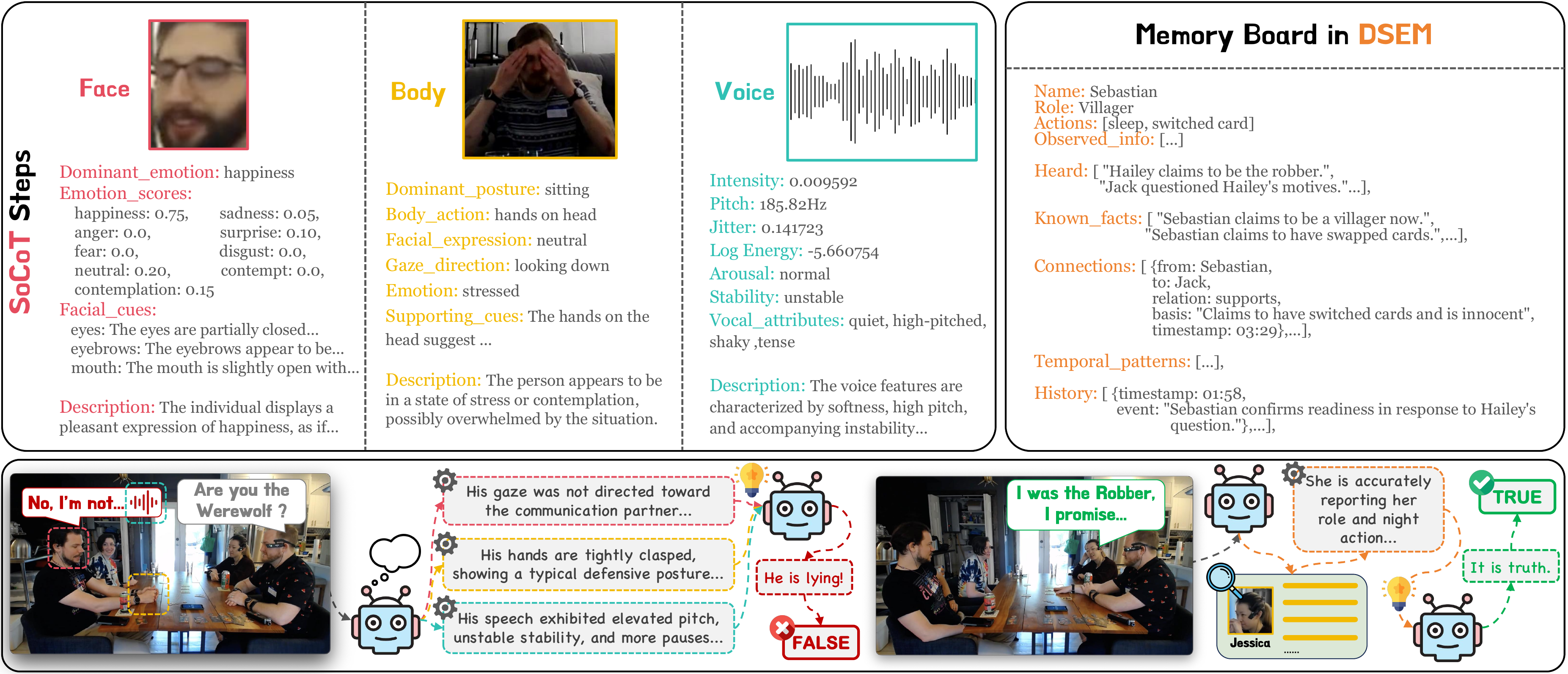}
   \caption{
   \textbf{Demonstration of SoCoT and DSEM Modules.} (Left) The SoCoT extracts multimodal behavioral primitives from Face, Body, and Voice. (Right) The DSEM module maintains a structured `Memory Board' tracking participants' evolving beliefs and social states. (Bottom) An example of SoCoT's and DSEM's reasoning steps. 
   }
    \label{fig:SoCoT_DSEM_demo}
     \vspace{-3ex}
\end{figure*}

\section{New Modules for Deception Reasoning}
\label{sec:method}

Reasoning about social deception requires grounding multimodal cues and tracking complex social states. To address this, we introduce two components: a Social Chain-of-Thought (SoCoT) reasoning pipeline and a Dynamic Social Epistemic Memory (DSEM) module. (Implementation details in Appendix~\ref{appendix:socot_pipeline} and~\ref{appendix:dsem_module}.)

\subsection{SoCoT: Social Chain-of-Thought}

SoCoT enforces a structured reasoning process that mirrors human social cognition, transforming raw multimodal inputs into a sequence of interpretable steps. 
It mitigates the inherent difficulty of MLLMs in achieving robust visual grounding, facilitating effective utilization of multimodal information.

\noindent\textbf{Step 1: Low-level Perception.}
We preprocess the Body, Face, and Voice inputs before reasoning. Speaker bounding boxes serve only as offline localization signals for visual cropping (with \textit{MTCNN}~\cite{zhang2016joint} applied to the face area), while \textit{parselmouth}~\cite{jadoul2018introducing} and \textit{librosa}~\cite{mcfee2015librosa} extract acoustic features.
SoCoT then requires MLLMs to perform explicit grounding to analyze the multimodal information first 
to extract \textit{behavioral primitives} (facial expressions, body gestures, and vocal cues)  as shown in Figure~\ref{fig:SoCoT_DSEM_demo}.
By forcing the model to describe these non-verbal cues before any reasoning occurs, it converts continuous audiovisual signals into symbolic, structured cues.

\noindent\textbf{Step 2: High-level Social Reasoning.}
Based on the extracted primitives, the model performs 
inference to approximate the speaker’s mental state and strategy. 
It simulates Theory-of-Mind reasoning to predict possible goals and social motives.

\noindent\textbf{Step 3: Decision and Rationale Generation.}
All intermediate inferences are aggregated to produce a final deception judgment along with an evidence-grounded justification. 
This provides interpretability and enables inspection of reasoning consistency.

\begin{table}[t]
\centering
\caption{Evaluations of closed-source models for the Persuasive Strategy Classification task. We report F1 scores for each of the six categories, along with the average F1 score and Joint Accuracy. (Open-source results in Appendix~\ref{appendix:additional_results}.)}
\vspace{-0.2cm}
\label{tab:strategy_results}
\setlength{\tabcolsep}{2pt}
\scalebox{0.65}{%
\begin{tabular}{l|l|cccccc}
\toprule
\rowcolor{cyan!20}
 & \textbf{Metric} &
\multicolumn{1}{c}{\textbf{\begin{tabular}{@{}c@{}}GPT4o\\-mini\end{tabular}}} &
\multicolumn{1}{c}{\textbf{GPT4o}} &
\multicolumn{1}{c}{\textbf{\begin{tabular}{@{}c@{}}GPT5\\-nano\end{tabular}}} &
\multicolumn{1}{c}{\textbf{\begin{tabular}{@{}c@{}}GPT5\\-mini\end{tabular}}} &
\multicolumn{1}{c}{\textbf{\begin{tabular}{@{}c@{}}Gemini\\2.5-pro\end{tabular}}} &
\multicolumn{1}{c}{\textbf{\begin{tabular}{@{}c@{}}Claude\\3.5-haiku\end{tabular}}} \\ \midrule
\multirow{8}{*}{\rotatebox{90}{Ego4D}} & Identity & \textbf{84.4} & 67.5 & 74.6 & 65.4 & 68.0 & 71.6 \\
 & Accusation & 45.1 & \textbf{47.6} & 40.0 & 40.0 & 45.6 & 44.4 \\
 & Defense & 22.2 & 29.9 & \textbf{36.5} & 31.9 & 34.3 & 36.0 \\
 & Evidence & 49.0 & 61.7 & 60.7 & 62.9 & \textbf{66.3} & 50.6 \\
 & Interrogation & 72.0 & 82.3 & 81.0 & 82.3 & \textbf{84.1} & 76.5 \\
 & Call for Action & 51.0 & \textbf{65.0} & 50.7 & 59.7 & 52.6 & 45.4 \\ \cmidrule{2-8}
 & Avg F1 Score & 53.9 & \textbf{59.0} & 57.3 & 57.0 & 58.5 & 54.1 \\
 & Joint Accuracy & 62.9 & 63.2 & 58.4 & \textbf{63.3} & 54.0 & 46.6 \\ \midrule
\multirow{8}{*}{\rotatebox{90}{YouTube}} & Identity & \textbf{72.6} & 53.7 & 60.4 & 52.5 & 50.0 & 68.7 \\
 & Accusation & 51.3 & \textbf{55.4} & 53.6 & 52.8 & 49.7 & 52.0 \\
 & Defense & 22.2 & 27.8 & 35.1 & 30.6 & 36.6 & \textbf{39.5} \\
 & Evidence & 46.9 & 52.8 & 54.0 & 53.8 & \textbf{54.6} & 44.2 \\
 & Interrogation & 75.0 & 80.1 & \textbf{85.9} & 84.5 & 85.3 & 73.8 \\
 & Call for Action & 62.5 & \textbf{72.3} & 59.0 & 59.1 & 60.8 & 55.2 \\ \cmidrule{2-8}
 & Avg F1 Score & 55.1 & 57.0 & \textbf{58.0} & 55.6 & 56.2 & 55.6 \\
 & Joint Accuracy & 55.5 & \textbf{57.3} & 55.6 & 54.0 & 49.8 & 47.4 \\
\bottomrule
\end{tabular}}
\vspace{-0.4cm}
\end{table}

\begin{table*}[!t]
\centering
\caption{
Evaluations on MIDA-Ego4D/MIDA-YouTube. Performance is reported using Overall Accuracy, Binary Accuracy (on $TRUE$/$FALSE$ samples), Macro-Averaged Precision, Recall, and F1-score, as well as fine-grained F1-scores for the $TRUE$, $FALSE$, and $NEUTRAL$ categories. (\colorbox{cyan!20}{Blue:} closed-source; \colorbox{yellow!20}{yellow:} open-source.)
}
\vspace{-1ex}
\label{tab:mida_closed_open}
\scalebox{0.65}{
\setlength{\tabcolsep}{3pt}
\begin{tabular}{l|l|cccccc||cccccc}
\toprule
\multicolumn{1}{c|}{\cellcolor{cyan!20}} 
  & 
  \multirow{1}{*}{\cellcolor{cyan!20}\textbf{Metric}}
  & \multicolumn{1}{c}{\cellcolor{cyan!20}\textbf{\begin{tabular}{@{}c@{}}GPT-4o\\mini\end{tabular}}}
  & \multicolumn{1}{c}{\cellcolor{cyan!20}\textbf{GPT-4o}}
  & \multicolumn{1}{c}{\cellcolor{cyan!20}\textbf{\begin{tabular}{@{}c@{}}GPT5\\nano\end{tabular}}}
  & \multicolumn{1}{c}{\cellcolor{cyan!20}\textbf{\begin{tabular}{@{}c@{}}GPT5\\mini\end{tabular}}}
  & \multicolumn{1}{c}{\cellcolor{cyan!20}\textbf{\begin{tabular}{@{}c@{}}Gemini\\2.5-pro\end{tabular}}}
  & \multicolumn{1}{c||}{\cellcolor{cyan!20}\textbf{\begin{tabular}{@{}c@{}}Claude\\3.5-haiku\end{tabular}}}
  & \multicolumn{1}{c}{\cellcolor{yellow!20}\textbf{\begin{tabular}{@{}c@{}}Llama\\3-8B\end{tabular}}}
  & \multicolumn{1}{c}{\cellcolor{yellow!20}\textbf{\begin{tabular}{@{}c@{}}InternVL\\3.5-8B\end{tabular}}}
  & \multicolumn{1}{c}{\cellcolor{yellow!20}\textbf{\begin{tabular}{@{}c@{}}Qwen2\\VL-7B\end{tabular}}}
  & \multicolumn{1}{c}{\cellcolor{yellow!20}\textbf{\begin{tabular}{@{}c@{}}Qwen2.5\\VL-7B\end{tabular}}}
  & \multicolumn{1}{c}{\cellcolor{yellow!20}\textbf{\begin{tabular}{@{}c@{}}Qwen3\\8B\end{tabular}}}
  & \multicolumn{1}{c}{\cellcolor{yellow!20}\textbf{\begin{tabular}{@{}c@{}}DeepSeek\\R1-8B\end{tabular}}} \\
\midrule

\multirow{8}{*}{\rotatebox{90}{Ego4D}} 
 & Accuracy          
 & 66.3 & \textbf{74.0} & 67.9 & 69.7 & 68.6 & 71.7
 & 51.2 & 64.3 & 65.3 & \textbf{76.4} & 67.1 & 39.3 \\

 & Accuracy (Binary) 
 & \textbf{39.4} & 27.6 & 14.2 & 10.2 & 26.0 & 4.7
 & \textbf{68.5} & 34.7 & 51.2 & 33.9 & 22.8 & 56.7 \\ \cmidrule{2-14}

 & Macro-Precision   
 & 43.6 & \textbf{52.7} & 39.1 & 38.4 & 43.2 & 42.0
 & 38.8 & 43.4 & 41.8 & \textbf{53.3} & 42.7 & 40.1 \\

 & Macro-Recall      
 & 47.3 & \textbf{52.5} & 37.9 & 36.3 & 46.3 & 35.6
 & 46.7 & 47.4 & \textbf{49.2} & 47.1 & 43.8 & 46.3 \\

 & Macro-F1          
 & 44.5 & \textbf{51.2} & 38.1 & 36.5 & 44.2 & 35.2
 & 34.2 & 43.8 & 43.7 & \textbf{48.8} & 42.6 & 32.4 \\ \cmidrule{2-14}

 & F1-TRUE           
 & \textbf{36.1} & 31.5 & 19.4 & 16.0 & 25.9 & 8.4
 & 33.6 & 33.9 & 43.6 & \textbf{48.9} & 26.1 & 31.7 \\

 & F1-FALSE          
 & 9.1 & \textbf{29.9} & 6.1 & 4.1 & 19.2 & 5.9
 & 0.0 & 10.7 & 0.0 & 0.0 & \textbf{12.7} & 7.4 \\

 & F1-NEUTRAL        
 & 88.3 & \textbf{92.4} & 88.7 & 89.5 & 87.4 & 91.5
 & 69.0 & 86.6 & 87.5 & \textbf{93.0} & 89.0 & 58.1 \\

\midrule \midrule
\multirow{8}{*}{\rotatebox{90}{YouTube}} 
 & Accuracy          
 & 65.9 & \textbf{69.8} & 63.8 & 63.2 & 64.9 & 62.1
 & 39.5 & 62.0 & 54.9 & 64.0 & 61.3 & 37.9 \\

 & Accuracy (Binary) 
 & \textbf{34.0} & 28.6 & 14.9 & 11.7 & 28.6 & 4.4
 & \textbf{45.2} & 35.2 & 20.5 & 12.7 & 14.4 & 42.5 \\ \cmidrule{2-14}

 & Macro-Precision   
 & 54.8 & \textbf{62.5} & 51.0 & 52.7 & 51.4 & 56.8
 & 43.8 & 49.3 & 34.1 & \textbf{53.4} & 47.5 & 40.5 \\

 & Macro-Recall      
 & \textbf{52.3} & 51.2 & 41.3 & 39.8 & 48.1 & 36.1
 & 44.1 & \textbf{50.7} & 40.4 & 40.7 & 40.4 & 41.0 \\

 & Macro-F1          
 & 53.4 & \textbf{54.2} & 42.4 & 40.4 & 49.4 & 33.8
 & 35.4 & \textbf{49.8} & 36.7 & 41.2 & 41.3 & 33.3 \\ \cmidrule{2-14}

 & F1-TRUE           
 & \textbf{38.4} & 32.3 & 27.4 & 21.8 & 31.3 & 4.4
 & 29.8 & \textbf{32.5} & 29.7 & 26.2 & 20.5 & 23.9 \\ 

 & F1-FALSE          
 & 34.9 & \textbf{41.6} & 14.8 & 14.2 & 33.1 & 12.0
 & 16.2 & \textbf{33.9} & 0.0 & 11.3 & 19.6 & 23.4 \\

 & F1-NEUTRAL        
 & 86.8 & \textbf{88.6} & 85.0 & 85.1 & 83.7 & 85.1
 & 60.2 & 83.1 & 80.3 & \textbf{86.1} & 83.9 & 52.6 \\

\bottomrule
\end{tabular}
}
\vspace{-1ex}
\end{table*}

\subsection{DSEM: Dynamic Social Epistemic Memory}

DSEM complements social reasoning by maintaining a structured, time-evolving representation of each participant’s beliefs and knowledge. 
For each player $p$, we maintain a structured \textit{Memory Board} $M_p^t$, 
containing key fields (\textit{Role}, \textit{Actions}, \textit{Observed Info}, \textit{Heard}, \textit{Known Facts}, \textit{Connections}, \textit{Reactions}, and \textit{Temporal Patterns}, \textit{History}) in a JSON-formatted file, which the MLLM is prompted to index before making a judgment and to update after each round.
We model this MLLM-driven update process as a state transition function, $f_{\text{DSEM}}$:
$$ M_p^{t+1} = f_{\text{DSEM}}(M_p^t, E_{t+1}, O_{t+1}) $$
where $M_p^t$ is the knowledge state at the previous step, $E_{t+1}$ is the new social event, and $O_{t+1}$ is the associated observation.
This converts unstructured dialogue into a dynamic memory board suitable for reasoning, and can be generalized to other multi-person social scenarios.

SoCoT and DSEM jointly form a modular architecture for social deception reasoning: 
SoCoT provides interpretable step-wise inference, while DSEM offers persistent and structured cognitive context. 
This design moves MLLMs toward more faithful modeling of human-like social intelligence within complex multi-party interactions.

\section{Experiments}
\label{sec:experiments}

We benchmark 12 MLLMs on the two MIDA tasks and ablate the contributions of temporal context, modality, and our proposed reasoning modules. Additional comparisons are deferred to the Appendix: open-source persuasive strategy results (App.~\ref{appendix:strategy_open}), classical PLM baselines~\cite{lai2023werewolf} (App.~\ref{appendix:strategy_comparison}), prior deception datasets~\cite{perez2015deception,zadeh2019boxoflies} (App.~\ref{appendix:mida_comparison}), and reasoning paradigms~\cite{wei2022chain,lewis2020retrieval,kosinski2023theory} (App.~\ref{appendix:baseline_modules}).

\vspace{-2ex}
\subsection{Experimental Setup}
\label{sec:models}

\noindent\textbf{Models.}
We benchmark 12 representative MLLMs from both proprietary and open-source families: GPT-4o~\cite{openai2024gpt4o}, GPT-4o-mini~\cite{openai2024gpt4omini}, GPT5-nano/mini~\cite{openai2025gpt5nano}, Gemini-2.5-pro~\cite{Google2025Gemini2_5}, Claude-3.5-haiku~\cite{Anthropic2024Claude3_5}, Llama-3-8B~\cite{grattafiori2024llama}, InternVL3.5-8B~\cite{wang2025internvl3}, Qwen2/2.5-VL-7B~\cite{wang2024qwen2,bai2025qwen2}, Qwen3-8B~\cite{yang2025qwen3}, and DeepSeek-R1-8B~\cite{guo2025deepseek}.

\noindent\textbf{Evaluation Metrics.}
Given the class imbalance (Sec.~\ref{sec:dataset_analysis}), we report \textit{F1-score} and \textit{Joint Accuracy} for Persuasive Strategy Classification, and \textit{Accuracy}, \textit{Binary Accuracy} on \texttt{TRUE}/\texttt{FALSE} samples, per-class \textit{F1-TRUE/FALSE/NEUTRAL}, and \textit{Macro-F1} for MIDA Deception Assessment.

\begin{table*}[t]
\centering
\caption{
Temporal Ablation Study on MIDA-Ego4D (GPT-4o-mini). 
(Left) Compares the model with full dialogue history against the prior dialogue removed (w/o History) model. (Right) Compares the Text-only model with 1-frame/3-frame vision models.
}
\vspace{-1ex}
\label{tab:temporal_effects}
\scalebox{0.63}
{%
\begin{tabular}{l|l|cC{2.2cm}||C{2.4cm}cc}
\toprule
\rowcolor{cyan!20}
\textbf{Task} & \textbf{Metric} & \textbf{Default w/ History} & \textbf{w/o History} & \textbf{Text-only} & \textbf{w/ Vision (1-frame)} & \textbf{w/ Vision (3-frame)} \\
\midrule
\multirow{8}{*}{Strategy} & Identity Declaration & \textbf{84.4} {(+0.0)} & 81.1 {\textcolor{red!70!black}{(-3.3)}} & \textbf{84.4} {(+0.0)} & 77.5 {\textcolor{red!70!black}{(-6.9)}} & 73.2 {\textcolor{red!70!black}{(-11.2)}} \\
 & Accusation & 45.1 {(+0.0)} & \textbf{47.3} {\textcolor{green!50!black}{(+2.2)}} & 45.1 {(+0.0)} & \textbf{49.4} {\textcolor{green!50!black}{(+4.3)}} & 46.6 {\textcolor{green!50!black}{(+1.5)}} \\
 & Defense & 22.2 {(+0.0)} & \textbf{22.7} {\textcolor{green!50!black}{(+0.5)}} & 22.2 {(+0.0)} & \textbf{23.3} {\textcolor{green!50!black}{(+1.1)}} & 18.2 {\textcolor{red!70!black}{(-4.0)}} \\
 & Evidence & 49.0 {(+0.0)} & \textbf{56.1} {\textcolor{green!50!black}{(+7.1)}} & \textbf{49.0} {(+0.0)} & 46.3 {\textcolor{red!70!black}{(-2.7)}} & 44.4 {\textcolor{red!70!black}{(-4.6)}} \\
 & Interrogation & 72.0 {(+0.0)} & \textbf{77.6} {\textcolor{green!50!black}{(+5.6)}} & \textbf{72.0} {(+0.0)} & 63.2 {\textcolor{red!70!black}{(-8.8)}} & 63.8 {\textcolor{red!70!black}{(-8.2)}} \\
 & Call for Action & 51.0 {(+0.0)} & \textbf{53.1} {\textcolor{green!50!black}{(+2.1)}} & \textbf{51.0} {(+0.0)} & 46.8 {\textcolor{red!70!black}{(-4.2)}} & 50.3 {\textcolor{red!70!black}{(-0.7)}} \\ \cmidrule{2-7}
 & Avg F1 Score & 53.9 {(+0.0)} & \textbf{56.3} {\textcolor{green!50!black}{(+2.4)}} & \textbf{53.9} {(+0.0)} & 51.1 {\textcolor{red!70!black}{(-2.8)}} & 49.4 {\textcolor{red!70!black}{(-4.5)}} \\
 & Joint Accuracy & \textbf{62.9} {(+0.0)} & 61.9 {\textcolor{red!70!black}{(-1.0)}} & \textbf{62.9} {(+0.0)} & 55.1 {\textcolor{red!70!black}{(-7.8)}} & 55.3 {\textcolor{red!70!black}{(-7.6)}} \\
\midrule 
\midrule
\multirow{5}{*}{MIDA} & Accuracy & 66.3 {(+0.0)} & \textbf{72.2} {\textcolor{green!50!black}{(+5.9)}} & 66.3 {(+0.0)} & \textbf{69.4} {\textcolor{green!50!black}{(+3.1)}} & 68.9 {\textcolor{green!50!black}{(+2.6)}} \\
 & Accuracy (Binary) & \textbf{39.4} {(+0.0)} & 13.4 {\textcolor{red!70!black}{(-26.0)}} & \textbf{39.4} {(+0.0)} & 38.6 {\textcolor{red!70!black}{(-0.8)}} & 35.4 {\textcolor{red!70!black}{(-4.0)}} \\ \cmidrule{2-7}
 & Macro-Precision & 43.6 {(+0.0)} & \textbf{46.4} {\textcolor{green!50!black}{(+2.8)}} & 43.6 {(+0.0)} & \textbf{46.1} {\textcolor{green!50!black}{(+2.5)}} & 45.6 {\textcolor{green!50!black}{(+2.0)}} \\
 & Macro-Recall & \textbf{47.3} {(+0.0)} & 42.3 {\textcolor{red!70!black}{(-5.0)}} & 47.3 {(+0.0)} & \textbf{48.9} {\textcolor{green!50!black}{(+1.6)}} & 48.7 {\textcolor{green!50!black}{(+1.4)}} \\
 & Macro-F1 & \textbf{44.5} {(+0.0)} & 42.1 {\textcolor{red!70!black}{(-2.4)}} & 44.5 {(+0.0)} & \textbf{47.0} {\textcolor{green!50!black}{(+2.5)}} & 46.6 {\textcolor{green!50!black}{(+2.1)}} \\
\bottomrule
\end{tabular}
}
\vspace{-1ex}
\end{table*}

\begin{figure*}[t]
  \centering
   \includegraphics[width=0.75\linewidth]{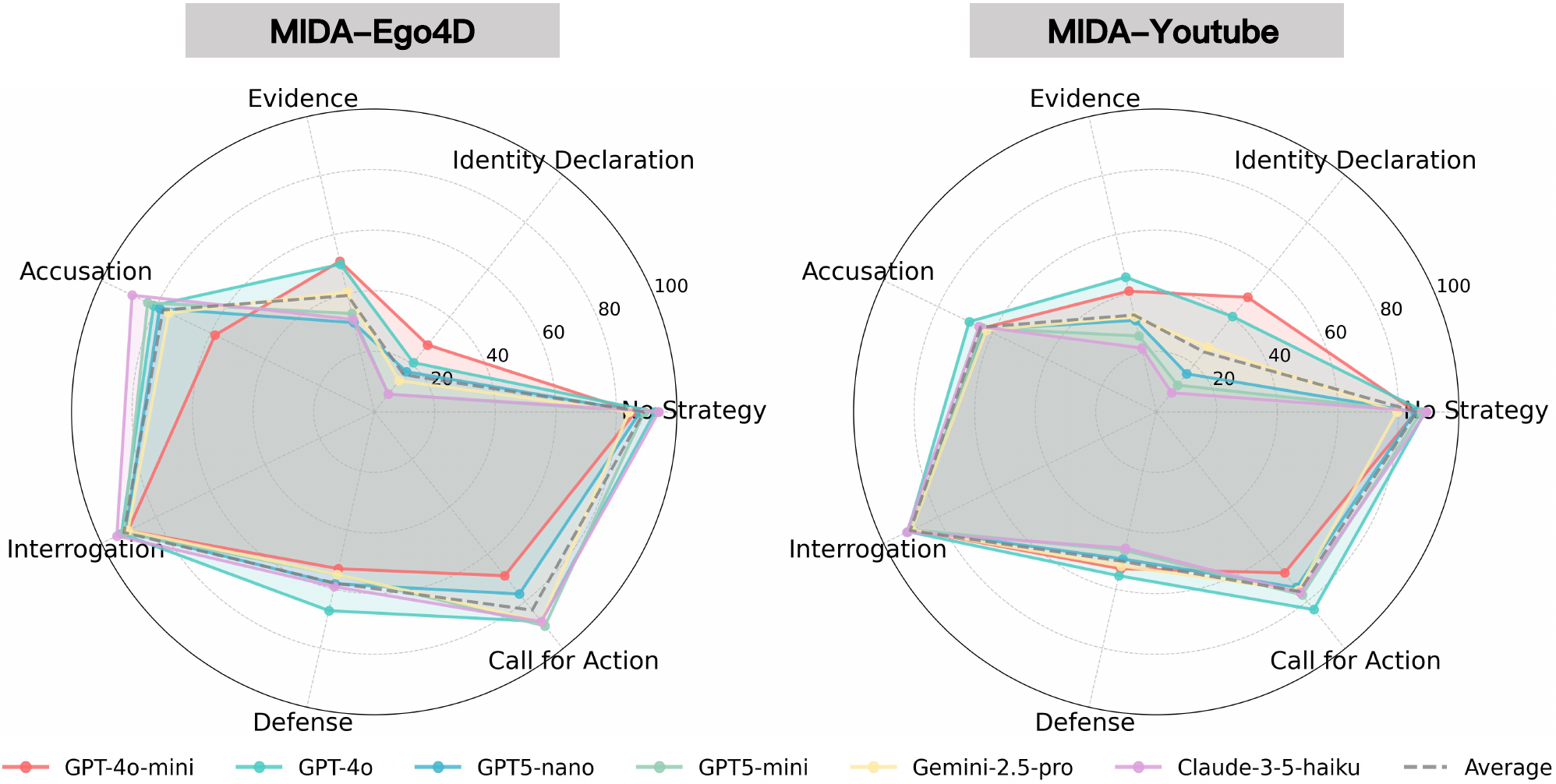}
   \caption{Radar chart of MLLMs’ accuracy in the MIDA task across Persuasive Strategy categories.}
    \label{fig:radar_chart}
    \vspace{-3ex}
\end{figure*}

\subsection{Main Benchmark Results}
\label{sec:benchmark_results}

\noindent\textbf{Performance on Persuasive Strategy Classification.}
Persuasive strategy classification is challenging, with no single dominant model across all conditions (Table~\ref{tab:strategy_results}).
On MIDA-Ego4D, closed-source MLLMs are closely grouped: GPT-4o attains the best average F1 (59.0) and competitive joint accuracy (63.2\%); GPT5-mini and Gemini-2.5-pro follow with average F1 of 57.0–58.5; and GPT5-nano leads on \textit{Defense} (36.5) with solid \textit{Interrogation} (81.0).
Models generally excel on structurally distinctive categories (\textit{Identity Declaration}, \textit{Interrogation}) but struggle with the nuanced, context-dependent language of \textit{Defense} and \textit{Call for Action}.

On the more deceptive and linguistically diverse MIDA-YouTube, performance drops slightly and the ranking becomes less uniform: GPT-5-nano achieves the highest average F1 (58.0) and leads \textit{Interrogation} (85.9), reflecting better generalization to spontaneous dialogue, while GPT-4o keeps the best joint accuracy (57.3\%). Across both datasets, \textit{Defense} remains the universal pain point, marking it as a key area for future work. Open-source results are in Appendix~\ref{appendix:strategy_open}.

\noindent\textbf{Performance on MIDA.}
On the core deception assessment task, results are highly uneven across both groups (Table~\ref{tab:mida_closed_open}).
Among \textbf{closed-source} models, GPT-4o leads overall on Ego4D, reflecting stronger holistic comprehension of multi-party discourse. However, the \textit{Binary Accuracy} metric, which isolates decisive \texttt{TRUE}/\texttt{FALSE} predictions, tells a different story: the smaller GPT-4o-mini attains the highest Binary accuracy (39.4 / 34.0 on Ego4D / YouTube) and the strongest F1-TRUE (36.1 / 38.4).

Among \textbf{open-source} models, the pattern persists at lower absolute levels. Qwen2.5-VL-7B achieves the best Macro-F1 on Ego4D (48.8) and competitive YouTube performance (41.2), driven by F1-TRUE (48.9) and F1-NEUTRAL (93.0); InternVL3.5-8B leads on YouTube Macro-F1 (49.8); and Llama-3-8B excels in Binary accuracy (68.5 / 45.2), again favoring decisive judgments at the cost of recall. 

This discrepancy exposes a shared pathology: (i) even the strongest models are heavily biased toward \texttt{NEUTRAL}, overestimating uncertainty as an artifact of conservative alignment; and (ii) none demonstrates a robust ``Theory of Mind''~\cite{kosinski2023theory}, collapsing on factual claims that require mental-state reasoning. Although GPT-4o and Qwen2.5-VL stand out in their respective groups, overall Macro-F1 underscores that MLLMs' social reasoning is still immature and far from reliable for real-world deception assessment.

\begin{table*}[!t]
\centering
\caption{
Evaluation of MLLMs with SoCoT and DSEM on the MIDA-Ego4D Dataset. Performance is reported using Overall Accuracy, Binary Accuracy, and Macro-Averaged Precision, Recall, and F1-score.
}
\label{tab:module_results_close}
\scalebox{0.84}{
\begin{tabular}{l|c|ccc|c}
\toprule
\rowcolor{cyan!20}
\textbf{Closed-source MLLMs} & \textbf{GPT-4o-mini} & \textbf{SoCoT-Face} & \textbf{SoCoT-Body} & \textbf{SoCoT-Voice} & \textbf{w/ DSEM} \\
\midrule
Accuracy
 & 66.3 (+0.0)
 & \textbf{74.3} \textcolor{green!50!black}{(+8.0)}
 & 71.5 \textcolor{green!50!black}{(+5.2)}
 & 70.7 \textcolor{green!50!black}{(+4.4)}
 & 70.7 \textcolor{green!50!black}{(+4.4)} \\
Accuracy (Binary)
 & 39.4 (+0.0)
 & 32.3 \textcolor{red!70!black}{(-7.1)}
 & 37.8 \textcolor{red!70!black}{(-1.6)}
 & 36.2 \textcolor{red!70!black}{(-3.2)}
 & \textbf{41.7} \textcolor{green!50!black}{(+2.3)} \\
\midrule
Macro-Precision
 & 43.6 (+0.0)
 & \textbf{47.9} \textcolor{green!50!black}{(+4.3)}
 & 46.5 \textcolor{green!50!black}{(+2.9)}
 & 45.8 \textcolor{green!50!black}{(+2.2)}
 & 46.5 \textcolor{green!50!black}{(+2.9)} \\
Macro-Recall
 & 47.3 (+0.0)
 & 46.4 \textcolor{red!70!black}{(-0.9)}
 & 48.3 \textcolor{green!50!black}{(+1.0)}
 & 47.6 \textcolor{green!50!black}{(+0.3)}
 & \textbf{50.2} \textcolor{green!50!black}{(+2.9)} \\
Macro-F1
 & 44.5 (+0.0)
 & 47.1 \textcolor{green!50!black}{(+2.6)}
 & 47.2 \textcolor{green!50!black}{(+2.7)}
 & 46.4 \textcolor{green!50!black}{(+1.9)}
 & \textbf{47.8} \textcolor{green!50!black}{(+3.3)} \\
\midrule
\rowcolor{yellow!20}
\textbf{Open-source MLLMs} & \textbf{Qwen2.5VL} & \textbf{SoCoT-Face} & \textbf{SoCoT-Body} & \textbf{SoCoT-Voice} & \textbf{w/ DSEM} \\
\midrule
Accuracy
 & 76.4 (+0.0)
 & 77.1 \textcolor{green!50!black}{(+0.7)}
 & 76.9 \textcolor{green!50!black}{(+0.5)}
 & 76.9 \textcolor{green!50!black}{(+0.5)}
 & \textbf{77.9} \textcolor{green!50!black}{(+1.5)} \\
Accuracy (Binary)
 & 33.9 (+0.0)
 & 27.6 \textcolor{red!70!black}{(-6.3)}
 & 29.1 \textcolor{red!70!black}{(-4.8)}
 & 30.7 \textcolor{red!70!black}{(-3.2)}
 & \textbf{36.2} \textcolor{green!50!black}{(+2.3)} \\
\midrule
Macro-Precision
 & 53.3 (+0.0)
 & 50.2 \textcolor{red!70!black}{(-3.1)}
 & 49.6 \textcolor{red!70!black}{(-3.7)}
 & 49.3 \textcolor{red!70!black}{(-4.0)}
 & \textbf{60.9} \textcolor{green!50!black}{(+7.6)} \\
Macro-Recall
 & 47.1 (+0.0)
 & 44.1 \textcolor{red!70!black}{(-3.0)}
 & 44.6 \textcolor{red!70!black}{(-2.5)}
 & 45.1 \textcolor{red!70!black}{(-2.0)}
 & \textbf{48.3} \textcolor{green!50!black}{(+1.2)} \\
Macro-F1
 & 48.8 (+0.0)
 & 45.8 \textcolor{red!70!black}{(-3.0)}
 & 46.1 \textcolor{red!70!black}{(-2.7)}
 & 46.5 \textcolor{red!70!black}{(-2.3)}
 & \textbf{50.4} \textcolor{green!50!black}{(+1.6)} \\
\bottomrule
\end{tabular}
}
\vspace{-0.3cm}
\end{table*}

\subsection{Ablation on Temporal Information}
\label{sec:ablation_temporal}

We investigate the role of temporal context by comparing GPT-4o-mini's performance under different conditions: (1) the default setting with full conversation history, (2) removing the text history (\textbf{w/o History}), and (3) providing multiple video frames (\textbf{1-frame vs. 3-frame}).

The results on MIDA-Ego4D, shown in Table~\ref{tab:temporal_effects}, reveal a crucial dichotomy. For strategy classification, removing textual history has minimal negative impact and sometimes even improves the Avg F1 score (+2.4). This suggests that identifying a statement's strategic function is largely a local phenomenon. In sharp contrast, removing the history causes a catastrophic drop in deception assessment performance (Accuracy (Binary) plummets from 39.4\% to 13.4\%). This demonstrates that assessing deception is a global task that relies on contextual reasoning over the entire conversation. 

Furthermore, the introduction of 1-frame vision demonstrates the essentiality of multimodal cues. For the MIDA task, incorporating 1-frame visual information yields a consistent improvement (Macro-F1 +2.5\%), confirming that social deception assessment requires grounding linguistic content in visual social cues. Conversely, the same visual input leads to a performance drop in strategy classification (Avg F1 -2.8\%), indicating that strategy classification remains a predominantly linguistic task where raw visual signals may act as distracting noise. While 3-frame input still provides gain over the text-only baseline for MIDA (+2.1\%), it shows a slight degradation compared to the 1-frame setting. 
This reinforces the conclusion that while multi-modal information is important, current MLLMs struggle to achieve robust visual grounding and distinguish salient social signals from noise when provided with increased temporal data.

\vspace{-1ex}
\subsection{MIDA across Persuasive Strategies}
\label{sec:analysis_categories}

Figure~\ref{fig:radar_chart} reveals a category-level imbalance: high accuracy on \textit{No Strategy} and \textit{Interrogation} is largely misleading, as these categories are dominated by \texttt{NEUTRAL} utterances. The true reasoning challenge lies in the information-rich categories \textit{Identity Declaration} (avg.\ 15.7\% on Ego4D, 25.3\% on YouTube) and \textit{Evidence} ($<$40\% on both), where performance collapses across all models. The GPT-4 family retains the clearest edge in these reasoning-intensive cases (e.g., GPT-4o-mini reaches 28.3\% / 48.4\% on \textit{Identity Declaration}), confirming that current MLLMs still struggle with the epistemic reasoning that governs truth and belief in social interactions.

\vspace{-1ex}
\subsection{Effect of Deception Reasoning Modules}
\label{sec:effect_soc_reasoning} 

We evaluate the proposed deception reasoning modules (Sec.~\ref{sec:method}) by instantiating them in two ways: (i) \textbf{SoCoT} with modality-specific cues (\textit{Face/Body/Voice}) and (ii) the \textbf{DSEM} memory module. Quantitative results are reported in Table~\ref{tab:module_results_close}; qualitative analysis is provided in Appendix~\ref{appendix:baseline_modules} (Figure~\ref{fig:rebuttalfig}). A comparison with reasoning paradigms, including Chain-of-Thought~\cite{wei2022chain}, Memory-Augmented RAG~\cite{lewis2020retrieval}, and ToM decomposition~\cite{kosinski2023theory}, is further detailed in Appendix~\ref{appendix:baseline_modules} and modality ablation studies in Appendix~\ref{appendix:modality_ablation}.

\noindent\textbf{Closed-source MLLMs.}
For GPT-4o-mini, each single-modality SoCoT variant improves overall Accuracy ($+4.4$ to $+8.0$) and Macro-F1 ($+1.9$ to $+2.7$), with SoCoT-Face attaining the highest Accuracy (74.3) and Macro-Precision (47.9). However, all three variants reduce Binary Accuracy ($-1.6$ to $-7.1$), so these holistic gains should not be interpreted as uniformly better discrimination between truth and falsehood. This trade-off indicates that current MLLMs can extract useful non-verbal descriptions yet still struggle to align them with factual and epistemic logic.
DSEM shows a different pattern, improving every reported metric over the GPT-4o-mini baseline and attaining the best Binary Accuracy (41.7; $+2.3$) and Macro-F1 (47.8; $+3.3$). This supports the value of a structured, time-evolving epistemic state for deception assessment.

\noindent\textbf{Open-source MLLMs.}
On Qwen2.5-VL, the text-only baseline reaches Acc 76.4, Binary 33.9, and Macro-F1 48.8.
SoCoT-Face/Body/Voice produce only small Accuracy gains ($+0.5$ to $+0.7$) while reducing Binary Accuracy ($-3.2$ to $-6.3$) and Macro-F1 ($-2.3$ to $-3.0$). Thus, for this 7B model, modality-specific descriptions do not translate into better deception reasoning.
DSEM, by contrast, improves every reported metric: Acc 77.9 ($+1.5$), Binary 36.2 ($+2.3$), Macro-F1 50.4 ($+1.6$), and Macro-Precision 60.9 ($+7.6$). The cross-model difference suggests that GPT-4o-mini's stronger vision and language alignment can partially exploit SoCoT features, whereas Qwen2.5-VL benefits more reliably from structured epistemic memory.

\section{Conclusion}
\label{sec:discussion_conclusion}
We introduced \textbf{MIDA}, a benchmark for assessing deception in multi-party social interactions, and comprehensively evaluated 12 state-of-the-art MLLMs on it. Our analysis identifies three core deficiencies: a lack of ``Theory of Mind'' for inferring hidden beliefs and intentions; an inability to ground language in salient social cues amid noise; and an overly conservative alignment that drives models toward safe, neutral answers. In short, current MLLMs are powerful \textit{knowledge engines} but not yet competent \textit{social agents}. DSEM consistently improves the reported metrics across both evaluated model families, while SoCoT yields mixed, model- and metric-dependent effects. Together, these results support epistemic memory and expose multimodal grounding as an important but unresolved direction.

\section*{Limitations}
While DSEM delivers consistent improvements and SoCoT reveals both gains and current limitations in multimodal grounding, MIDA remains a demanding benchmark, and several promising directions stand out for future work:
(i) \textit{context-adaptive alignment} strategies that are less risk-averse in adversarial, multi-party settings;
(ii) new architectures with \textit{integrated Theory-of-Mind reasoning} that track participants' evolving beliefs and intentions;
(iii) more robust methods for \textit{grounding language in non-verbal visual and acoustic cues}.
Addressing these challenges is a necessary step toward the truly perceptive and trustworthy AI systems required for seamless human--AI collaboration.

\section*{Ethical Considerations}
\textbf{Data sourcing.} MIDA is built on top of two existing corpora: Ego4D~\cite{grauman2022ego4d}, whose participants provided informed consent under the original release agreement, and publicly available YouTube recordings of recreational \emph{One Night Ultimate Werewolf} matches. We add only veracity labels and night-action annotations; no new video or audio is recorded, and no personally identifying information beyond what is already public is introduced.

\noindent\textbf{Dual-use and intended use.} Deception detection has well-documented dual-use risks, including surveillance, employment screening, and law-enforcement misuse~\cite{vrij2010pitfalls}. MIDA is explicitly framed as a \emph{diagnostic benchmark for the social-reasoning capabilities of MLLMs}, not as a deployable lie detector. The verifiable ground truth available here arises from the rules of a closed, gamified setting and \emph{does not transfer to real-world deception}, where labels are rarely objective. We therefore caution against using MIDA-trained models in high-stakes applications such as forensic interrogation or hiring.

\noindent\textbf{Annotation.} Night-action annotation was carried out by the authors and collaborators with full knowledge of the project goals; no crowdworkers were employed for sensitive labelling.

\section*{Acknowledgments}
This work was supported by JSPS KAKENHI Grant Number JP24K02956 and JST ASPIRE Grant Number JPMJAP2303.

\bibliography{main}

\clearpage
\appendix

\begin{center}
   {\Large \bf Supplementary Material \par}
\end{center}
\vspace{1.5em}

\etocdepthtag.toc{mtappendix}
\etocsettagdepth{mtmain}{none}
\etocsettagdepth{mtappendix}{subsection}
\renewcommand{\contentsname}{Appendix Contents}
\setcounter{tocdepth}{2}
\tableofcontents
\clearpage

\numberwithin{equation}{section}
\numberwithin{figure}{section}
\numberwithin{table}{section}

\section{Generalizability of MIDA}
\label{sec:generalizability}

MIDA is a diagnostic benchmark grounded in the game-based setting of Werewolf, which provides unusually clear truth conditions for studying social reasoning. Semantic game-state metadata is used to establish ground-truth labels but is withheld from models at inference; speaker bounding boxes are used only for offline visual cropping. These labels are objective under the game rules, while transfer to real-world deception remains to be validated.

\noindent\textbf{(1) Reliance on Game Structure.} Semantic game-state metadata is used for verifiable ground-truth labeling during dataset construction and is not exposed to models during inference. The proposed SoCoT and DSEM are designed as general-purpose modules for broader social reasoning tasks.

\noindent\textbf{(2) Why Werewolf?} It serves as a verifiable playground to solve the ground-truth scarcity in deception reasoning. Obtaining deception labels in any multi-party interactions is extremely challenging. Existing monologue-style deception datasets like \textit{Real-Life Trial} and \textit{Box of Lies} are also domain-constrained.

\section{Additional Benchmark Results.}
\label{appendix:additional_results}

\subsection{Performance on Persuasive Strategy (Open-Source)}
\label{appendix:strategy_open}
Open-source MLLMs show diverse but complementary strengths across the six persuasive categories, reflecting varying levels of pragmatic reasoning and discourse awareness (Table~\ref{tab:strategy_results_open}).

On the Ego4D benchmark, InternVL3.5-8B delivers the highest average F1 (55.2), suggesting a well-balanced understanding of strategic discourse. Qwen3-8B follows closely (54.2) yet achieves the best joint accuracy (70.0\%), indicating more coherent multi-label prediction. Examining category-level results reveals a clear specialization pattern: Qwen3-8B excels in linguistically explicit and visually grounded strategies such as \textit{Identity Declaration}, \textit{Evidence}, and \textit{Call for Action}, while InternVL3.5-8B dominates in \textit{Interrogation} with an F1 of 86.1. In contrast, implicit and context-dependent behaviors, \textit{Defense} and \textit{Accusation}, remain challenging across the board, with only DeepSeek-R1-8B showing partial improvement.

On the YouTube dataset, which contains spontaneous, colloquial interactions, all models experience moderate degradation. Both InternVL3.5-8B and Qwen3-8B achieve the best average F1 (56.6), though Qwen3-8B again attains the top joint accuracy (61.1\%). Its strong results on \textit{Interrogation} (80.8) and \textit{Call for Action} (73.3) suggest better adaptability to unstructured dialogue. Meanwhile, DeepSeek-R1-8B continues to show robust recognition of adversarial intentions, particularly in \textit{Defense} (35.4) and \textit{Accusation} (46.9).

Across both corpora, open-source models approach the performance tier of closed-source MLLMs on linguistically salient or visually grounded categories but remain brittle when persuasion depends on subtle context and implicit social goals. These results point to a persistent gap between linguistic pattern recognition and genuine pragmatic reasoning in current open-source multimodal LLMs.

\begin{table*}[t]
\centering
\caption{Evaluations of open-source models for the Persuasive Strategy Classification task on the Ego4D and YouTube. We report F1 scores for each of the six categories, along with the average F1 score and Joint Accuracy (all six labels predicted correctly). 
}
\label{tab:strategy_results_open}
\scalebox{0.65}
{%
\begin{tabular}{l|l|cccccc}
\toprule
\rowcolor{yellow!20}
\textbf{Dataset}         & \textbf{Models}      & \textbf{Llama-3-8B} & \textbf{InternVL3\_5-8B} & \textbf{Qwen2-VL-7B} & \textbf{Qwen2.5-VL-7B} & \textbf{Qwen3-8B} & \textbf{DeepSeek-R1-8B} \\
 \midrule

\multirow{8}{*}{Ego4D}   & Identity Declaration & 70.2                & 78.6                     & 50.3                 & 64.8                   & \textbf{82.5}     & 58.8                    \\
                         & Accusation           & 26.5                & 34.3                     & 36.4                 & 8.3                    & 27.6              & \textbf{43.2}           \\
                         & Defense              & 3.9                 & 24.1                     & 14.9                 & 3.9                    & 11.0              & \textbf{25.6}           \\
                         & Evidence             & 39.9                & 51.0                     & 40.9                 & 13.3                   & \textbf{52.8}     & 46.3                    \\
                         & Interrogation        & 60.4                & \textbf{86.1}            & 37.8                 & 75.4                   & 82.1              & 74.6                    \\
                         & Call for Action      & 53.6                & 56.9                     & 46.9                 & 50.0                   & \textbf{69.2}     & 56.9                    \\ \cmidrule{2-8}
                         & Avg F1 Score         & 42.4                & \textbf{55.2}            & 37.9                 & 35.9                   & 54.2              & 50.9                    \\
                         & Joint Accuracy       & 49.0                & 60.9                     & 29.1                 & 67.6                   & \textbf{70.0}     & 53.9                    \\ \midrule
\multirow{8}{*}{YouTube} & Identity Declaration & 44.6                & 67.5                     & 39.0                 & 55.7                   & \textbf{75.5}     & 46.6                    \\
                         & Accusation           & 42.8                & \textbf{47.7}            & 42.0                 & 21.9                   & 42.7              & 46.9                    \\
                         & Defense              & 9.4                 & 33.2                     & 23.0                 & 12.4                   & 18.8              & \textbf{35.4}           \\
                         & Evidence             & 33.2                & 48.4                     & 38.1                 & 15.8                   & \textbf{48.4}     & 36.5                    \\
                         & Interrogation        & 56.2                & 78.0                     & 43.8                 & 63.6                   & \textbf{80.8}     & 74.0                    \\
                         & Call for Action      & 61.4                & 65.0                     & 58.6                 & 59.5                   & \textbf{73.3}     & 64.7                    \\ \cmidrule{2-8}
                         & Avg F1 Score         & 41.3                & 56.6                     & 40.8                 & 38.2                   & \textbf{56.6}     & 50.7                    \\
                         & Joint Accuracy       & 41.3                & 53.3                     & 27.0                 & 56.7                   & \textbf{61.1}     & 44.1    \\      
\bottomrule
\end{tabular}
}
\end{table*}

\begin{table*}[t]
\centering
\caption{Comparison of open- and closed-source LLMs (0-shot) with the method in ~\cite{lai2023werewolf} on the Persuasive Strategy Classification Task on Ego4D. ``+C'' denotes Modeling with Context Embedding (prior utterances). 
}
\label{tab:strategy_results_comparison_1}
\scalebox{0.56}
{%
\begin{tabular}{l|ccc|ccc|cc|cc}
\toprule
\rowcolor{Gray!20}
\textbf{Metric}      & \textbf{BERT} & \textbf{RoBERTa} & \textbf{MT-BERT} & \textbf{BERT + C} & \textbf{RoBERTa + C} & \textbf{MT-BERT + C} & \textbf{GPT-4o-mini} & \textbf{GPT-4o} & \textbf{Qwen2.5-VL-7B} & \textbf{Qwen3-8B} \\
 \midrule
Identity & 82.6          & 81.7             & 80.9             & 79.9              & 81.5                 & 79.8                 & \textbf{84.4}        & 67.5            & 64.8                   & 82.5              \\
Accusation           & 48.8          & 51.7             & 51.5             & 52                & \textbf{59.4}        & 54.4                 & 45.1                 & 47.6            & 8.3                    & 27.6              \\
Defense              & 29.3          & 33.1             & 25.9             & 33.8              & 33                   & \textbf{36.5}        & 22.2                 & 29.9            & 3.9                    & 11.0              \\
Evidence             & 54.2          & 51.7             & 53.6             & 57.1              & 52.4                 & 61.5                 & 49.0                 & \textbf{61.7}   & 13.3                   & 52.8              \\
Interrogation        & 82.8          & 83.4             & 83               & 81                & \textbf{83.5}        & 83.2                 & 72.0                 & 82.3            & 75.4                   & 82.1              \\
Call for Action      & 39.4          & 43.3             & 56.6             & 49.5              & 43.7                 & 50.8                 & 51.0                 & 65.0            & 50.0                   & \textbf{69.2}     \\
\midrule
Avg F1 Score         & 56.2          & 57.5             & 58.6             & 58.9              & 58.9                 & \textbf{61.0}        & 53.9                 & 59.0            & 35.9                   & 54.2              \\
Joint Accuracy       & 65.1          & 63.4             & 65.5             & 65                & 64.6                 & 66.3                 & 62.9                 & 63.2            & 67.6                   & \textbf{70.0}   \\
\bottomrule
\end{tabular}
}
\end{table*}

\begin{table*}[!t]
\centering
\caption{Comparison of open- and closed-source LLMs (0-shot) with the method in ~\cite{lai2023werewolf} on the Persuasive Strategy Classification Task on Ego4D. ``+C'' denotes Modeling with Context Embedding (prior utterances), and ``+V'' denotes Modeling with Video Representation. 
}
\label{tab:strategy_results_comparison_2}
\scalebox{0.54}
{%
\begin{tabular}{l|ccc|ccc|cc|cc}
\toprule
\rowcolor{Gray!20}
\textbf{Metric}      & \textbf{BERT + V} & \textbf{RoBERTa + V} & \textbf{MT-BERT + V}   &\textbf{B. + C + V}  &\textbf{Ro. + C + V} &\textbf{MT. + C + V} & \textbf{GPT-4o-mini} & \textbf{GPT-4o} & \textbf{Qwen2.5-VL-7B} & \textbf{Qwen3-8B} \\

 \midrule
Identity & 81.5     & 79.8        & 79.9          & \textbf{84.5} & 82.7            & 80.7            & 84.4                 & 67.5            & 64.8                   & 82.5              \\
Accusation           & 52.1     & 51.4        & 51.9          & 52.8          & \textbf{58.5}   & 55.2            & 45.1                 & 47.6            & 8.3                    & 27.6              \\
Defense              & 28.4     & 31.3        & 35.4          & 34.5          & 35.4            & \textbf{36.1}   & 22.2                 & 29.9            & 3.9                    & 11.0              \\
Evidence             & 52.8     & 54.6        & 53.3          & 54.9          & 53.4            & 60.5            & 49.0                 & \textbf{61.7}   & 13.3                   & 52.8              \\
Interrogation        & 83.3     & 82.8        & \textbf{84.8} & 82.7          & 83.8            & 83.6            & 72.0                 & 82.3            & 75.4                   & 82.1              \\
Call for Action      & 42.4     & 50.1        & 53.9          & 47.3          & 46.1            & 50.0            & 51.0                 & 65.0            & 50.0                   & \textbf{69.2}     \\
\midrule
Avg F1 Score         & 56.7     & 58.3        & 59.8          & 59.4          & 60.0            & \textbf{61.0}   & 53.9                 & 59.0            & 35.9                   & 54.2              \\
Joint Accuracy       & 64.5     & 64.0        & 62.1          & 66.5          & 66.1            & 66.3            & 62.9                 & 63.2            & 67.6                   & \textbf{70.0}    \\
\bottomrule
\end{tabular}
}
\end{table*}

\begin{figure*}[t]
  \centering
   \includegraphics[width=0.99\linewidth]{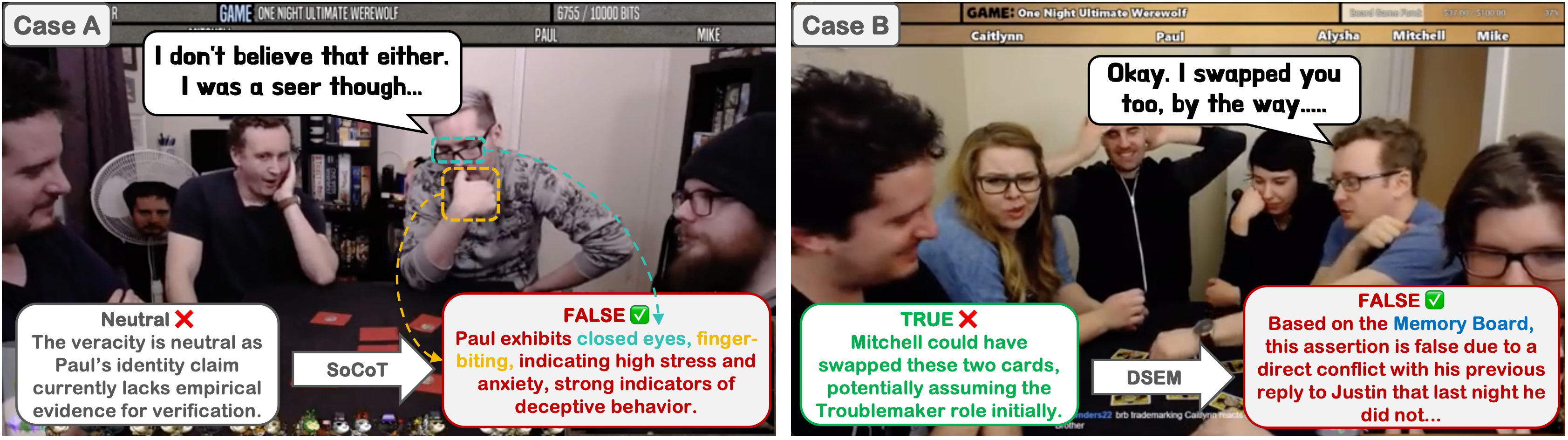}
  \caption{\textbf{Qualitative analysis of failure modes and the corrective effects of SoCoT and DSEM. }
   \textbf{(Case A)} illustrates a perception failure where the baseline relies on text; SoCoT corrects this by extracting multimodal stress indicators. \textbf{(Case B)} demonstrates a cognitive failure resolved by DSEM, which tracks evolving facts via the \textit{Memory Board} to detect logical inconsistencies.
   }
    \label{fig:rebuttalfig}
\end{figure*}

\subsection{Comparative Analysis on Persuasive Strategy}
\label{appendix:strategy_comparison}

Tables~\ref{tab:strategy_results_comparison_1} and \ref{tab:strategy_results_comparison_2} together present a comprehensive comparison across three modeling paradigms: (1) classical PLM baselines from~\cite{lai2023werewolf}, (2) their multi-modal variants with visual and contextual fusion, and (3) modern closed- and open-source multimodal LLMs. The results illustrate how these architectures differ in their ability to capture discourse-level persuasive strategies.

\noindent\textbf{PLM baselines.}
From Table~\ref{tab:strategy_results_comparison_1}, classical transformer models (BERT, RoBERTa, MT-BERT) gain modest improvements when augmented with contextual utterances (\texttt{+C}). Average F1 increases from roughly 56–58 to a maximum of 61.0, indicating that shallow context modeling helps maintain discourse coherence but remains limited to local linguistic cues. These models perform reliably on structurally distinctive strategies such as \textit{Identity Declaration} and \textit{Interrogation}, but struggle with more implicit, intention-driven categories such as \textit{Accusation} and \textit{Defense}.

\noindent\textbf{Multi-modal variants with visual/context fusion.}
Table~\ref{tab:strategy_results_comparison_2} shows that adding video signals (\texttt{+V}) and combining them with contextual embeddings (\texttt{+C+V}) yields more consistent gains. MT-BERT+V reaches 59.8 average F1, while MT-BERT+C+V achieves 61.0, the strongest among classical architectures. These improvements are especially pronounced on visually grounded strategies such as \textit{Identity Declaration} and \textit{Interrogation}. However, even with explicit multi-modal fusion, performance gains on subtle social-intent categories remain limited, implying that handcrafted fusion still cannot fully capture pragmatic inference.

\noindent\textbf{Closed- and open-source LLMs.}
Closed-source systems display more robust and generalizable reasoning. \textbf{GPT-4o} matches the best C+V baselines in average F1 (59.0) and clearly surpasses them on context-heavy strategies such as \textit{Evidence} (61.7) and \textit{Call for Action} (65.0). \textbf{GPT-4o-mini}, though lighter, maintains competitive performance (53.9 average F1) and excels on \textit{Identity Declaration} (84.4), suggesting strong pragmatic priors even under smaller capacity.

Open-source vision-language models show a different trade-off: while \textbf{Qwen2.5-VL-7B} underperforms in average F1 (35.9), it achieves high multi-label consistency (67.6\% joint accuracy). \textbf{Qwen3-8B} offers the best balance, 54.2 average F1 and the highest joint accuracy overall (70.0\%), driven by strong results on visually grounded strategies (e.g., 69.2 on \textit{Call for Action}). However, both open-source models continue to struggle on strategies involving implicit social inference, particularly \textit{Defense} and \textit{Accusation}.

Together, Tables~\ref{tab:strategy_results_comparison_1} and \ref{tab:strategy_results_comparison_2} reveal complementary strengths across modeling paradigms: traditional PLMs and their C/V-enhanced variants capture local and visually grounded signals effectively, whereas LLM-based multimodal models leverage broader discourse and pragmatic context. Closed-source models currently provide the strongest overall generalization, while the best open-source systems are rapidly closing the gap, especially in multi-label structural consistency, pointing toward emerging parity in open-source persuasive reasoning.

\begin{table*}[!t]
\centering
\caption{
Comparison of open- and closed-source LLMs with methods from Real-Life Trial~\cite{perez2015deception} and Box of Lies~\cite{zadeh2019boxoflies} on the MIDA-Ego4D Dataset. Performance is reported using Overall Accuracy, Binary Accuracy (on $TRUE$/$FALSE$ samples), Macro-Averaged Precision, Recall, and F1-score, as well as fine-grained F1-scores for the $TRUE$, $FALSE$, and $NEUTRAL$ categories.
}
\label{tab:MIDA_results}
\scalebox{0.73}
{%
\begin{tabular}{l|cc|cc|cc}
\toprule
\rowcolor{Gray!20}
\textbf{Models}   & \textbf{Real-Life Trial} & \textbf{Box of Lies} & \textbf{GPT-4o-mini} & \textbf{GPT-4o} & \textbf{Qwen2.5-VL-7B} & \textbf{Qwen3-8B} \\ \midrule
Accuracy          & 63.8            & 68.6        & 66.3                 & 74.0            & \textbf{76.4}          & 67.1              \\
Accuracy (Binary) & 11.8            & 7.1         & \textbf{39.4}        & 27.6            & 33.9                   & 22.8              \\ \midrule
Macro-Precision   & 32.3            & 40.0        & 43.6                 & 52.7            & \textbf{53.3}          & 42.7              \\
Macro-Recall      & 32.8            & 35.5        & 47.3                 & \textbf{52.5}   & 47.1                   & 43.8              \\
Macro-F1          & 32.5            & 35.5        & 44.5                 & \textbf{51.2}   & 48.8                   & 42.6              \\
\midrule
F1-TRUE           & 13.5            & 12.6        & 36.1                 & 31.5            & \textbf{48.9}          & 26.1              \\
F1-FALSE          & 0.0             & 3.7         & 9.1                  & \textbf{29.9}   & 0.0                    & 12.7              \\
F1-NEUTRAL        & 84.1            & 90.1        & 88.3                 & 92.4            & \textbf{93.0}          & 89.0                                   \\     
\bottomrule
\end{tabular}
}
\end{table*}

\subsection{Comparative Analysis on MIDA.}
\label{appendix:mida_comparison}
Table~\ref{tab:MIDA_results} compares classical multimodal deception-detection systems from prior work~\cite{perez2015deception, zadeh2019boxoflies} with both closed- and open-source multimodal LLMs on the MIDA-Ego4D benchmark. The contrast reveals clear differences in robustness, decision reliability, and category-level biases.

Traditional deception detectors such as Real-Life Trial and Box of Lies achieve moderate overall accuracy (63.8–68.6) but exhibit extremely poor binary discrimination on factual claims, with binary accuracies below 12\%. Their \texttt{TRUE} and \texttt{FALSE} F1-scores remain near zero, indicating an overreliance on ``neutral'' judgments and limited capacity to infer speaker intent—consistent with earlier findings that hand-engineered multimodal cues struggle with conversational deception.

Closed-source multimodal LLMs significantly outperform these baselines. GPT-4o achieves the best overall balance, obtaining 51.2 Macro-F1 and strong F1 on \texttt{FALSE} (29.9), reflecting genuine discriminative ability rather than neutral defaulting. GPT-4o-mini, despite its smaller capacity, attains the highest binary accuracy (39.4) among closed-source models, suggesting a more decisive prediction pattern on factual claims. Both models deliver high \texttt{NEUTRAL} F1 ($\geq$ 88), but unlike traditional baselines, they maintain non-trivial performance on \texttt{TRUE}/\texttt{FALSE}, indicating more calibrated alignment between factual reasoning and social context.

The open-source models show complementary strengths. Qwen2.5-VL-7B achieves the highest overall accuracy (76.4) and the best \texttt{TRUE} F1 (48.9), but collapses entirely on \texttt{FALSE} (0.0), revealing a strong bias toward one-sided factual interpretation. Qwen3-8B is more balanced, outperforming classical baselines across all metrics and achieving a reasonable \texttt{FALSE} F1 (12.7), yet still trailing GPT-4o in overall consistency. Both open-source models maintain strong \texttt{NEUTRAL} F1 (89–93), further highlighting the general tendency of current MLLMs to over-predict neutrality when uncertain.

Overall, the results indicate that while modern LLMs, both closed and open, dramatically outperform classical deception-detection architectures, they still struggle with fine-grained factual discrimination. High \texttt{NEUTRAL} performance masks a persistent difficulty in assigning confident \texttt{TRUE}/\texttt{FALSE} labels, underscoring an open challenge in designing models that can infer latent beliefs, intentions, and subtle inconsistencies in spontaneous dialogue.

For reference, Table~\ref{tab:floor_baselines} reports two uninformative floor baselines. The majority-class baseline obtains high Overall Accuracy by always predicting \texttt{NEUTRAL}, but has zero Binary Accuracy; the uniform-random baseline remains near chance. This contrast motivates considering Overall Accuracy together with Binary Accuracy and Macro-F1.

\begin{table}[t]
\centering
\caption{Floor baselines on MIDA-Ego4D.}
\label{tab:floor_baselines}
{\small
\setlength{\tabcolsep}{4pt}
\begin{tabular}{lccc}
\toprule
\textbf{Baseline} & \textbf{Accuracy} & \textbf{Binary} & \textbf{Macro-F1} \\
\midrule
Majority-class & 84.5 & 0.0 & 30.5 \\
Uniform-random & 33.5 & 33.1 & 23.7 \\
\bottomrule
\end{tabular}
}
\end{table}

\subsection{Comparison with Baseline Reasoning Modules.}
\label{appendix:baseline_modules}
To further benchmark the efficacy of the proposed framework, SoCoT and DSEM are compared against several established reasoning paradigms: standard Chain-of-Thought (CoT)~\cite{wei2022chain}, Memory-Augmented RAG~\cite{lewis2020retrieval}, and Theory-of-Mind (ToM) decomposition~\cite{kosinski2023theory}. While CoT and Memo-RAG demonstrate competitive performance in general accuracy, they often fail to capture the nuanced, interactive nature of multi-party deception. In contrast, SoCoT and DSEM yield superior results in the social reasoning context by specifically grounding inferences in multimodal cues and persistent epistemic states. These results validate the proposed modules as a more effective paradigm for modeling complex social interactions.

\begin{table*}[t]
\centering
\caption{Comparison of SoCoT and DSEM against baseline reasoning paradigms on MIDA-Ego4D: CoT~\cite{wei2022chain}, Memo-RAG~\cite{lewis2020retrieval}, and ToM-Decomp~\cite{kosinski2023theory}.}
\label{tab:baselines_comparison}
\vspace{1mm}
\scalebox{0.86}{
\begin{tabular}{l|c|ccc|cc}
\toprule
\rowcolor{Gray!20}
\textbf{Metric} & \textbf{GPT-4o-mini} & \textbf{CoT} & \textbf{Memo-RAG} & \textbf{ToM-Decomp} & \textbf{SoCoT} & \textbf{DSEM} \\
\midrule
Accuracy & 66.3 & 65.9 & 70.2 & 44.2 & \textbf{72.2} & 70.7 \\
Accuracy (Binary) & 39.4 & 40.9 & 29.9 & 12.6 & 35.4 & \textbf{41.7} \\
Macro-F1 & 44.5 & 45.9 & 44.3 & 28.5 & 46.9 & \textbf{47.8} \\
\bottomrule
\end{tabular}
}
\end{table*}

\section{Modality Ablation Study}
\label{appendix:modality_ablation}

\begin{table*}[t]
\centering
\caption{Modality Ablation Study on MIDA-Ego4D (GPT-4o-mini). All models are compared against the \textbf{Text-only} baseline. Values are shown as \emph{score ($\Delta$ vs baseline, p.p.)}.}
\label{tab:modality_ablation}
\scalebox{0.61}
{%
\begin{tabular}{l|c|ccc|ccc|c}
\toprule
\rowcolor{cyan!20}
\textbf{Models} & \textbf{Text-only} & \textbf{Face \& Body} & \textbf{Face \& Voice} & \textbf{Body \& Voice} & \textbf{Vision \& DSEM} & \textbf{Face \& DSEM} & \textbf{Voice \& DSEM} & \textbf{All Modality} \\ \midrule
Accuracy & 66.3 {(+0.0)} & 59.1 {\textcolor{red!70!black}{(-7.2)}} & 59.4 {\textcolor{red!70!black}{(-6.9)}} & 56.6 {\textcolor{red!70!black}{(-9.7)}} & 69.4 {\textcolor{green!50!black}{(+3.1)}} & 70.2 {\textcolor{green!50!black}{(+3.9)}} & 69.9 {\textcolor{green!50!black}{(+3.6)}} & 54.8 {\textcolor{red!70!black}{(-11.5)}} \\
Accuracy (Binary) & 39.4 {(+0.0)} & 26.0 {\textcolor{red!70!black}{(-13.4)}} & 27.6 {\textcolor{red!70!black}{(-11.8)}} & 33.9 {\textcolor{red!70!black}{(-5.5)}} & 29.9 {\textcolor{red!70!black}{(-9.5)}} & 25.2 {\textcolor{red!70!black}{(-14.2)}} & 27.6 {\textcolor{red!70!black}{(-11.8)}} & 15.0 {\textcolor{red!70!black}{(-24.4)}} \\ \midrule
Macro-Precision & 43.6 {(+0.0)} & 36.2 {\textcolor{red!70!black}{(-7.4)}} & 38.1 {\textcolor{red!70!black}{(-5.5)}} & 36.6 {\textcolor{red!70!black}{(-7.0)}} & 42.8 {\textcolor{red!70!black}{(-0.8)}} & 44.1 {\textcolor{green!50!black}{(+0.5)}} & 45.0 {\textcolor{green!50!black}{(+1.4)}} & 33.2 {\textcolor{red!70!black}{(-10.4)}} \\
Macro-Recall & 47.3 {(+0.0)} & 37.1 {\textcolor{red!70!black}{(-10.2)}} & 39.9 {\textcolor{red!70!black}{(-7.4)}} & 38.7 {\textcolor{red!70!black}{(-8.6)}} & 43.1 {\textcolor{red!70!black}{(-4.2)}} & 43.8 {\textcolor{red!70!black}{(-3.5)}} & 44.5 {\textcolor{red!70!black}{(-2.8)}} & 30.8 {\textcolor{red!70!black}{(-16.5)}} \\
Macro-F1 & 44.5 {(+0.0)} & 35.7 {\textcolor{red!70!black}{(-8.8)}} & 38.3 {\textcolor{red!70!black}{(-6.2)}} & 35.7 {\textcolor{red!70!black}{(-8.8)}} & 42.9 {\textcolor{red!70!black}{(-1.6)}} & 43.8 {\textcolor{red!70!black}{(-0.7)}} & 44.4 {\textcolor{red!70!black}{(-0.1)}} & 31.4 {\textcolor{red!70!black}{(-13.1)}} \\
\bottomrule
\end{tabular}
}
\end{table*}

We conducted a modality ablation study to understand the true contribution of non-verbal information, with results shown in Table \ref{tab:modality_ablation}. The findings are profoundly counter-intuitive. Contrary to the assumption that additional modalities would aid social reasoning, the data clearly shows that providing more sensory information to the baseline model significantly degrades performance.

Any combination of non-verbal modalities without our memory module (e.g., \textit{Face \& Body}, \textit{Face \& Voice}) results in a worse Macro-F1 score than the \textit{Text-only} baseline. The most damning evidence comes from the \textit{All Modality} condition. When the model is provided with all available information, its performance does not merely dip; it collapses. The overall Macro-F1 plummets from 44.5 to 31.4 (a 13.1-point drop), and more critically, the \textit{Accuracy (Binary)}---the model's core ability to distinguish truth from lies, experiences a catastrophic failure, falling from 39.4\% to 15.0\% (a massive 24.4 percentage point decrease).

Even when our \textit{DSEM} module is introduced, which slightly improves overall accuracy, the core \textit{Accuracy (Binary)} metric still suffers significantly (e.g., a 14.2 p.p. drop for \textit{Face \& Memory}). This strongly suggests that the current MLLM is unable to effectively utilize these additional modalities. Instead of providing clarifying social signals, the non-verbal data appears to act as overwhelming noise, confounding the model's reasoning process and leading to a chaotic and worse outcome. This highlights that effective multimodal fusion and grounding for high-level social reasoning is a critical, unsolved problem.

\section{Human Validation Study}
\label{appendix:human_validation}

\begin{table*}[t]
\centering
\caption{
Comparison of Model Performance on the MIDA-Ego4D Subset and the Human-validated Subset. Results for closed-source and open-source models are presented alongside the human-validated subset results for comparative analysis.
}
\label{tab:human_validation}
\scalebox{0.72}
{%
\begin{tabular}{l|cc|cc}
\toprule
\rowcolor{gray!20}
\textbf{Models}            & \textbf{GPT-4o-mini} & \textbf{Human-validated} & \textbf{Qwen2.5VL} & \textbf{Human-validated} \\
\midrule
Accuracy          & 57.9        & 57.9             & 76.9      & 76.9             \\
Accuracy (Binary) & 22.2        & 25.6             & 48.9      & 53.5             \\
\midrule
Macro-Precision   & 37.8        & 39.0             & 59.3      & 51.0             \\
Macro-Recall      & 34.5        & 35.0             & 54.2      & 51.0             \\
Macro-F1          & 36.1        & 36.9             & 56.1      & 51.0  \\
\bottomrule
\end{tabular}
}
\end{table*}

To verify the reliability of our semi-automatic annotation pipeline and ensure that our conclusions are not driven by dataset noise, we conduct a human validation study on a randomly sampled 20\% subset of the MIDA-Ego4D data. Human annotators checked and revised this subset, and model predictions on the same subset were compared against both the original labels and the corrected human-validated labels. The agreement between the original labels and the human-corrected labels is high (96.6\%), confirming the overall consistency of the dataset.

Table~\ref{tab:human_validation} reports model performance on (1) the original 20\% subset and (2) the same subset after human correction. For \textbf{GPT-4o-mini}, performance remains largely unchanged: overall accuracy stays at 57.9, while binary accuracy increases slightly from 22.2 to 25.6, and macro-level metrics improve marginally (e.g., Macro-F1 from 36.1 to 36.9). These small shifts indicate that GPT-4o-mini's errors on the full dataset are not primarily caused by annotation inconsistencies; instead, the model struggles with intrinsic reasoning challenges in the MIDA task.

In contrast, \textbf{Qwen2.5-VL-7B} shows a more notable gap before and after human correction. Although overall accuracy remains the same (76.9), binary accuracy increases from 48.9 to 53.5, suggesting that the model aligns more closely with human judgments when ambiguous factual claims are clarified. However, its macro metrics decrease (e.g., Macro-F1 from 56.1 to 51.0), revealing that part of its strong performance on the original subset may arise from exploiting annotation biases or majority-label tendencies.

Together, these results demonstrate that the evaluation is robust and fair:
(1) the human-validated subset shows extremely high agreement with the original annotations (96.6\%), indicating that the dataset contains very little label noise; and
(2) Model performance changes minimally after correction, confirming that our reported metrics are not artifacts of annotation errors. Although Qwen2.5-VL exhibits some fluctuations across metrics, these shifts reflect sensitivity to boundary cases rather than systematic bias in the dataset. Overall, the human validation study affirms that the MIDA benchmark provides a reliable and stable basis for model comparison.

\section{Compute Resources}
\label{appendix:compute_resources}

We summarize below the compute resources used to produce the results reported in Sec.~\ref{sec:experiments} and the appendix. All experiments are evaluation-only (zero-shot inference); no model training or fine-tuning is performed.

\noindent\textbf{Open-source MLLMs (local GPU inference).}
The six open-source models---Llama-3-8B, InternVL3.5-8B, Qwen2-VL-7B, Qwen2.5-VL-7B, Qwen3-8B, and DeepSeek-R1-8B---are run on a single NVIDIA A100 40GB GPU using HuggingFace Transformers with bfloat16 precision. Peak GPU memory usage is approximately 25--30~GB per model. Each model processes the full 2{,}360-utterance MIDA test set in roughly 10--15~hours, depending on whether visual frames are included. The main benchmark (Table~\ref{tab:mida_closed_open}, open-source columns) therefore consumed \textbf{$\sim$70 A100-hours}. The SoCoT/DSEM ablations on Qwen2.5-VL (Table~\ref{tab:module_results_close}, four variants on MIDA-Ego4D) added another \textbf{$\sim$15 A100-hours}, for a reported total of \textbf{$\sim$85 A100-hours} on a single A100 40GB.

\noindent\textbf{Closed-source MLLMs (API inference).}
GPT-4o, GPT-4o-mini, GPT-5-nano, GPT-5-mini, Gemini-2.5-Pro, and Claude-3.5-Haiku were accessed through their official APIs between August 2025 and January 2026. Each model was queried once per utterance with \texttt{temperature}$=0$. The main benchmark plus the SoCoT/DSEM, temporal, and modality ablations together comprise roughly 60{,}000 API calls, with cumulative spending of approximately \textbf{$\sim$300 USD}. Table~\ref{tab:api-snapshots} reports the requested identifiers and the served snapshots where they were observable; unpinned aliases are reported with their run window rather than assigned an unverifiable snapshot.

\begin{table*}[t]
\centering
\caption{Closed-source API identifiers used for the reported results. $^{\dagger}$Snapshot confirmed from API response metadata. $^{\ddagger}$Alias not pinned; served by the provider default from August 2025 to January 2026.}
\label{tab:api-snapshots}
{\small
\setlength{\tabcolsep}{8pt}
\begin{tabular}{lll}
\toprule
\textbf{Model} & \textbf{Requested identifier} & \textbf{Served snapshot} \\
\midrule
GPT-4o-mini      & \texttt{gpt-4o-mini}                  & \texttt{gpt-4o-mini-2024-07-18}$^{\dagger}$ \\
GPT-4o           & \texttt{gpt-4o}                       & Provider default (run window)$^{\ddagger}$ \\
GPT-5-nano       & \texttt{gpt-5-nano}                   & Provider default (run window)$^{\ddagger}$ \\
GPT-5-mini       & \texttt{gpt-5-mini}                   & Provider default (run window)$^{\ddagger}$ \\
Gemini-2.5-Pro   & \texttt{gemini-2.5-pro-exp-03-25}     & \texttt{gemini-2.5-pro-exp-03-25}$^{\dagger}$ \\
Claude-3.5-Haiku & \texttt{claude-3-5-haiku-20241022}   & \texttt{claude-3-5-haiku-20241022} \\
\bottomrule
\end{tabular}
}
\end{table*}

\noindent\textbf{Annotation pipeline.}
The semi-automated annotation pipeline (Sec.~\ref{sec:dataset}) uses Gemini-2.5-Pro to label all 2{,}360 utterances against the manually annotated night actions and game metadata, completing in roughly \textbf{$\sim$6 hours} of API time at an additional cost of approximately \textbf{50 USD}.

\noindent\textbf{Other compute.}
Preliminary experiments, prompt iteration, and failed runs not reported in the paper required additional compute on the same hardware. We estimate the cumulative budget across the full project at roughly \textbf{1.5--2$\times$} the figures reported above ($\sim$150 A100-hours and $\sim$500 USD in total API spending).

\section{Details for MIDA Annotation and Evaluation}
\label{appendix:annotation_eval}

\subsection{Details for MIDA Annotation}
\label{appendix:annotation_details}

In this section, we briefly summarize the annotation protocol behind MIDA and how it connects to the prompts shown in Appendix Figures (Game Rules -- Part 1 Figure~\ref{fig:appendix_GameRules_1}/Part 2 Figure~\ref{fig:appendix_GameRules_2}, Game Annotation -- Part 1 Figure~\ref{fig:appendix_annotation_1}/Part 2 Figure~\ref{fig:appendix_annotation_2}).

First, the \emph{Game Rules} prompts (Game Rules -- Part 1 and Part 2) provide annotators and LLMs with a complete, self-contained description of \textit{One Night Ultimate Werewolf}. These pages specify the role set, faction objectives, night action order, and win conditions. Crucially, they highlight that a player's initial and final roles may differ due to night actions such as Robber and Troublemaker, and that win conditions are determined by the \emph{final} card. This rule-based structure is what enables MIDA to define verifiable ground truth for every utterance in the dialogue.

Building on this foundation, the \emph{Game Annotation} prompts (Game Annotation -- Part 1 and Part 2) formalize how veracity labels are assigned at the utterance level. They define the three labels (\texttt{TRUE}, \texttt{FALSE}, \texttt{NEUTRAL}) in terms of the speaker's private knowledge state at the time of speaking, and introduce key principles such as the ``Outdated Information'' rule (statements remain TRUE if they were correct when acquired) and the ``Falsifiability'' principle (guesses, opinions, and inherently unverifiable statements must be labeled NEUTRAL). These pages also describe a suggested workflow: reconstructing the night actions to obtain a ``god's-eye'' view of the game, then performing line-by-line judgment with explicit justifications.

Finally, the prompts specify the required output format for both humans and LLM annotators: a structured table (or JSON) with timestamps, speaker, utterance, veracity label, and a short explanation that links each label to the game facts and the speaker's knowledge. Together, these components ensure that MIDA's annotations are grounded in deterministic game mechanics rather than subjective impression, and that evaluation of MLLMs on MIDA is transparent and reproducible.

\begin{figure*}[!htbp]
\begingroup
\scriptsize
\begin{tcolorbox}[
    colback=black!5!white, colframe=black, coltitle=white, colbacktitle=black!75!gray,
    fonttitle=\bfseries, title=Prompt for MIDA Annotation (Game Rules - Part 1),
    center title,
    arc=2mm, boxrule=0.5mm
]
\subsection*{Game Overview}
One Night Ultimate Werewolf is a social deduction game where players secretly take on the roles of either Villagers, Werewolves, or special characters. The goal is to identify and execute the Werewolves. The game is extremely fast-paced, consisting of a single night phase and a single day phase.

\tcbline
\subsection*{Game Setup}
\begin{itemize}
\item Card Selection: The number of game cards used should be three more than the number of players. 
\item Dealing: Cards are shuffled face down, and each player is dealt one card.
\item Center Cards: The three remaining cards, along with tokens that match all roles in play, are placed face down in the center of the table.
\item Viewing Roles: Each player secretly looks at their own card and places it face down in front of them.
\end{itemize}

\tcbline
\subsection*{Faction Objectives}
\textbf{Werewolf Team}
\begin{itemize}
    \item Werewolf: The Werewolves win if no Werewolves are executed. If all players survive, or if the only player executed is not a Werewolf, the Werewolf wins.
    \item Minion: The Minion wins if at least one Werewolf survives. If the Minion successfully protects all Werewolves and a good role (like a Villager or Hunter) is executed, the Minion wins. If all Werewolves are executed (including those who became Werewolves via the Robber or Troublemaker), the Minion loses.
\end{itemize}
\textbf{Village Team}
\begin{itemize}
    \item Villager, Seer, Robber, Troublemaker, Drunk, Insomniac, Hunter: The Villagers win if they successfully execute at least one Werewolf. 
\end{itemize}
\textbf{Unique Roles}
\begin{itemize}
\item Tanner: The Tanner wins if they are executed. This is the Tanner's only objective. They will pretend to be a Villager or a Werewolf, doing everything they can to convince other players to vote for them.
\item Doppelgänger: The Doppelgänger's win condition depends on the role they copy.
  (1) If they copy a Village team role (e.g., Seer), their win condition becomes "execute a Werewolf."
  (2) If they copy a Werewolf, their win condition becomes "no Werewolves are executed."
  (3) If they copy the Tanner, their win condition becomes "get executed."
\end{itemize}

\tcbline

\textbf{Key Rule 1:} Because of the Robber and Troublemaker, a player's initial role and final role may change during the night. Therefore, each player's final win condition is determined by the card they hold at the end of the game. This makes the game's deception, deduction, and betrayal more complex.

\end{tcolorbox}
\endgroup
\vspace{-2ex}
\captionof{figure}{The complete and detailed annotation prompt used to generate the MIDA dataset's veracity labels.(Game Rules - Part 1)}
\label{fig:appendix_GameRules_1}
\end{figure*}

\begin{figure*}[h!]
\begingroup
\scriptsize

\begin{tcolorbox}[
    colback=black!5!white, colframe=black, coltitle=white, colbacktitle=black!75!gray,
    fonttitle=\bfseries, title=Prompt for MIDA Annotation (Game Rules - Part 2),
    center title,
    arc=2mm, boxrule=0.5mm
]

\subsection*{The Night Phase}
During the night phase, all players close their eyes. A designated player, the Announcer, calls out each role in a specific order, silently counting to ten to allow that role to perform their night action. Roles without a special night action (such as the Villager, Tanner, and Hunter) do not wake up at night.

\textbf{Night Action Order:}
\begin{itemize}
\item Doppelgänger: Wakes up first and looks at another player's face-down card, immediately gaining that role's identity and abilities. If they copy a role with a night action, the Doppelgänger immediately performs that action.
\item Werewolves: Wake up and secretly identify their fellow Werewolves. If there is only one Werewolf, they can look at one of the center cards.
\item Minion: The Minion wakes up, and the Werewolves extend their thumbs to let the Minion know who they are.
\item Masons: Masons wake up and confirm each other's identities.
\item Seer: The Seer wakes up and can choose to look at another player's card or at two of the center cards.
\item Robber: The Robber wakes up and can swap their card with another player's card, then look at their new card.
\item Troublemaker: The Troublemaker wakes up and can swap the cards of two other players without looking at them.
\item Drunk: The Drunk wakes up and must exchange their card with one of the center cards without looking at it.
\item Insomniac: The Insomniac wakes up last and looks at their card to confirm whether their role was changed during the night.
\end{itemize}

\tcbline

\subsection*{The Day Phase}
After players wake up, they begin a discussion to deduce the Werewolves.
\begin{itemize}
\item Players are free to speak, but they cannot show their card to anyone.
\item Due to the actions of certain roles, a player's final identity may be different from their initially dealt card.
\end{itemize}

\tcbline

\subsection*{Voting and Game End}
After a few minutes of discussion, all players simultaneously vote by pointing.

\textbf{Execution:} The player with the most votes is executed and their role card is revealed. In case of a tie, all players who received the most votes are executed.

\textbf{Win Conditions:}
\begin{itemize}
  \item Village Team Wins: If at least one Werewolf is executed. The Village team still wins even if innocent players are executed alongside a Werewolf. The Village team also wins if there are no Werewolves in the game and no one is executed.
  \item Werewolf Team Wins: If at least one player is a Werewolf and no Werewolves are executed.
  \item Tanner Wins: The Tanner only wins if they are executed. The Tanner's victory is independent of the Werewolf or Village team's victory.
\end{itemize}

\end{tcolorbox}
\endgroup
\vspace{-2ex}
\captionof{figure}{The complete and detailed annotation prompt used to generate the MIDA dataset's veracity labels. (Game Rules - Part 2)}
\label{fig:appendix_GameRules_2}
\vspace{-2ex}
\end{figure*}

\begin{figure*}[t!]
\begingroup
\scriptsize
\begin{tcolorbox}[
    colback=black!5!white, colframe=black, coltitle=white, colbacktitle=black!75!gray,
    fonttitle=\bfseries, title=Prompt for MIDA Annotation (Game Annotation - Part 1),
    center title,
    arc=2mm, boxrule=0.5mm
]

\subsection*{Role: Game Analyst for AI Deception Detection}
\textbf{Objective:} Your task is to generate a high-precision ``Ground Truth" dataset for assessing deception. You will analyze game data and a full dialogue transcript from a single game and label each statement with its Truthfulness. \textbf{The core principle of your analysis is:} all judgments must be strictly based on the private information the speaker possessed at the moment they made the statement.

\textbf{Inputs:}
\begin{itemize}
    \item \textbf{Game State Data (JSON):}  (1) Contains decisive information such as the list of players, each player's starting and final role, and the voting outcome, and the exact nightActions that occurred. (2)  Interpretation of the votingOutcome array: The array index represents the voter (in the order of playerNames), and the number in the array represents the person who was voted for (e.g., the first number, 2, means player 0 "Justin" voted for player 2 "Paul").
    \item \textbf{Game Dialogue Transcript (CSV):} A complete record including timestamps, speaker, and dialogue content. (Note: The data may be spliced from two video segments, so the timer resets once.)
\end{itemize}
\tcbline
\subsection*{Core Task \& Labeling Rules:}
You need to analyze the dialogue transcript line by line and add a new Truthfulness label to each statement. The labels must adhere to the following definitions:

\textbf{TRUE:}
\begin{itemize}
    \item \textbf{Definition:} The speaker knows with certainty that what they are saying aligns with the facts they currently possess.
    \item \textbf{Applicable Scenarios:} (1) Honestly stating one's own role (e.g., a Villager whose card was not swapped says, "I'm a Villager").
    (2) Accurately describing a night action they performed (e.g., a Seer says, "I looked at the two center cards, and they were the Villager and the Robber").
    (3) Making a correct logical statement based on one's own certain information (e.g., an Insomniac confirms their role hasn't changed and says, "I'm still me").
\end{itemize}
\textbf{FALSE:}
\begin{itemize}
    \item \textbf{Definition:} The speaker knows with certainty that what they are saying contradicts the facts they currently possess, i.e., they are actively lying.
    \item \textbf{Applicable Scenarios:} (1) Lying about one's role or actions (e.g., a Werewolf says, "I am a Villager").
    (2) Fabricating false information to mislead others (e.g., a Villager falsely claims to be the Seer and checked someone).
    (3) Knowingly making a false accusation (e.g., Werewolf A accuses a good player to protect Werewolf B).
\end{itemize}
\textbf{NEUTRAL:}
\begin{itemize}
    \item \textbf{Definition:} The truthfulness of the statement cannot be determined based on the speaker's known information.
    \item \textbf{Applicable Scenarios:} (1) Asking questions, joking, or making irrelevant small talk.
    (2) Expressing subjective guesses or opinions (e.g., "I feel like Laura is suspicious," or "I think we should vote for Mitchell").
    (3) Proposing a possibility or strategic hypothesis (e.g., "If Justin is a Werewolf, then who would the other Werewolf be?").
\end{itemize}
\tcbline
\subsection*{Key Rules 2: I. Core Principles for Factual Statements}
These rules apply to any statement that makes a claim about a past or present game fact.

\textbf{Rule 1: Statements Aligning with Known Facts are TRUE.}
\begin{itemize}
\item Explanation: If a speaker makes a statement that is consistent with their private knowledge of the game state, it is labeled TRUE.
\item Example: A Villager who was not swapped by any night action says, "I am a Villager." This is TRUE.
\item Sub-rule 1a (The "Outdated Information" Rule): This includes statements based on information that was factually correct when acquired, even if the game state has since changed unbeknownst to the speaker. For instance, a Seer views Player A's card and sees "Villager." If Player A is later swapped, the Seer's statement "I saw Player A and they are a Villager" is still TRUE because it accurately reflects the information they possess.
\end{itemize}

\end{tcolorbox}
\endgroup
    \vspace{-2ex}
\captionof{figure}{The complete and detailed annotation prompt used to generate the MIDA dataset's veracity labels. (Game Annotation - Part 1)}
\label{fig:appendix_annotation_1}
\end{figure*}

\begin{figure*}[t!]
\begingroup
\scriptsize
\begin{tcolorbox}[
    colback=black!5!white, colframe=black, coltitle=white, colbacktitle=black!75!gray,
    fonttitle=\bfseries, title=Prompt for MIDA Annotation (Game Annotation - Part 2),
    center title,
    arc=2mm, boxrule=0.5mm
]

\textbf{Rule 2: Statements Contradicting Known Facts are FALSE.}
\begin{itemize}
\item Explanation: If a speaker makes a statement that they know to be inconsistent with their private knowledge, it is an intentional lie and is labeled FALSE. Example: A Werewolf says, "I am the Seer." This is FALSE.
\item Sub-rule 2a (The "Failed Lie" Rule): This applies even if the lie accidentally corresponds to some other unknown truth in the game. For instance, a Werewolf lies and claims "I am the Robber" to hide their identity. Even if another player's card is the Robber, the Werewolf's statement is FALSE because their intent was to misrepresent their own known reality.
\end{itemize}

\tcbline

\subsection*{Key Rules 2: II. The Falsifiability Principle for Non-Factual Statements}

This principle serves as the final adjudication rule for statements that are not direct claims of past or present fact. To maintain objectivity, if a statement's truthfulness cannot be fact-checked against the speaker's known reality, it must be labeled Neutral, regardless of any perceived deceptive intent.

\textbf{Rule 3: Statements Lacking a Factual Basis are Neutral.}
\begin{itemize}
\item Explanation: A statement made without sufficient information is a guess, opinion, or hypothesis.
\item Sub-rule 3a (The "Lucky Guess" Rule): A guess that happens to be correct is still Neutral. For example, a Villager with no information randomly accuses Player A of being a Werewolf. Even if Player A is the Werewolf, the statement is Neutral.
\item Sub-rule 3b (The "Overgeneralization" Rule): Extrapolating from incomplete information is also Neutral. For example, a lone Werewolf sees one "Villager" card in the center and claims, "The center cards are all safe." This is an overgeneralization and therefore Neutral.
\item Sub-rule 3c (The "Strategic Opinion" Rule): A subjective opinion ("I think," "I feel") is Neutral, even if its purpose is to mislead. For example, a Werewolf says, "I don't think Kaelan is a Werewolf" to protect their partner. This is a strategic opinion, not a factual lie, and is therefore Neutral.
\end{itemize}

\textbf{Rule 4: Inherently Unverifiable Statements are Neutral.}
\begin{itemize}
\item Explanation: Certain types of statements are impossible to verify at the moment they are uttered.
\item Sub-rule 4a (The "Future Intent" Rule): Statements about a future action ("I will vote for...") are Neutral because intent can change.
\item Sub-rule 4b (The "Hypothetical" Rule): Conditional statements ("If... then...") used for deduction are Neutral.
\item Sub-rule 4c (Questions and Commands): All questions and direct commands are inherently non-factual and are therefore Neutral.

\end{itemize}

\tcbline

\subsection*{Suggested Workflow}
\begin{enumerate}
    \item \textbf{Reconstruct and Document Night Events:} Your first and most critical task is to use the provided nightActions data, in conjunction with the strict Order of Night Actions, to deterministically reconstruct the step-by-step events of the night. The nightActions data is the ground truth of what happened. This summary must detail each action and the resulting change in card positions and player knowledge.
    \item \textbf{Line-by-Line Judgment:} With this ``god's-eye view" from the global analysis, read the dialogue transcript line by line. For each statement, ask yourself: "What were the 'facts' in the speaker's mind when they said this? Is this statement based on their known facts, a lie, or a guess?"
    \item \textbf{Assign Labels:} Assign a \texttt{TRUE}, \texttt{FALSE}, or \texttt{NEUTRAL} label and provide an explanation.
\end{enumerate}

\tcbline

\subsection*{Final Output Format}
Your final output should be a CSV table, including at least the following core columns:
\begin{itemize}
    \item \textbf{Timestamp} / \textbf{Speaker} / \textbf{Line} / \textbf{Truthfulness} (TRUE, FALSE, or NEUTRAL)
    \item \textbf{Explanation} (a concise explanation for the assigned Truthfulness label, connecting it to game facts and the speaker's knowledge). 
\end{itemize}

\end{tcolorbox}
\endgroup
    \vspace{-2ex}
\captionof{figure}{The complete and detailed annotation prompt used to generate the MIDA dataset's veracity labels.(Game Annotation - Part 2)}
\label{fig:appendix_annotation_2}
\end{figure*}

\begin{figure*}[h!]
\begingroup
\scriptsize
\begin{tcolorbox}[
    colback=black!5!white, colframe=black, coltitle=white, colbacktitle=black!75!gray,
    fonttitle=\bfseries, title=Prompt for MIDA Evaluation (Main Task - Part 1),
    center title,
    arc=2mm, boxrule=0.5mm
]

\subsection*{Given Role}
You are an expert analyst for the game "One Night Ultimate Werewolf".
Your task is to analyze a player's statement based on the provided game rules and the conversation history.

\tcbline
\subsection*{One Night Ultimate Werewolf: Core Rules}

\begin{enumerate}
    \item \textbf{Core Objective}
    \begin{itemize}
        \item \textbf{Village Team Wins:} If at least one Werewolf is executed.
        \item \textbf{Werewolf Team Wins:} If no Werewolves are executed (and at least one Werewolf is in play).
        \item \textbf{Tanner Wins:} If the Tanner is executed, the Tanner wins.
    \end{itemize}

    \item \textbf{Key Principles}
    \begin{itemize}
        \item \textbf{Final Role Matters:} A player's role can change during the night. The final win condition is determined by the card a player holds at the END of the game, not the start.
        \item \textbf{Information is Private:} A player only knows what their own role's action allowed them to see or do. Everything else is deduction.
        \item \textbf{Deception is Allowed:} Any player can lie about their role or actions.
    \end{itemize}

\tcbline

    \item \textbf{Role Descriptions by Faction}

    \textbf{Werewolf Team}
    \begin{itemize}
        \item \textbf{Werewolf:} At night, wakes up to see fellow Werewolves. A lone Werewolf may view one center card. Wins if no Werewolves are executed.
        \item \textbf{Minion:} At night, wakes up to see who the Werewolves are. The Werewolves do not know the Minion. Wins if at least one Werewolf survives (Minion is not executed). 
    \end{itemize}

    \textbf{Village Team}
    \begin{itemize}
        \item \textbf{Villager:} No special night actions. Goal is to help execute a Werewolf.
        \item \textbf{Seer:} At night, may look at one other player's card or at two of the center cards.
        \item \textbf{Robber:} At night, may swap their card with another player's card, and then looks at their new card.
        \item \textbf{Troublemaker:} At night, may swap the cards of two other players without looking at them.
        \item \textbf{Drunk:} At night, must exchange their card with one of the center cards without looking at it.
        \item \textbf{Insomniac:} Wakes up at the end of the night phase to look at their own card to see if their role was changed.
        \item \textbf{Hunter:} If the Hunter is executed, the player they are pointing at is also executed.
        \item \textbf{Mason:} At night, wakes up to see the other Mason (if one is in play).
    \end{itemize}

    \textbf{Unique Roles}
    \begin{itemize}
        \item \textbf{Tanner:} Is on their own team. Wins only if they are executed. They are not on the Village or Werewolf team.
        \item \textbf{Doppelgänger:} At night, looks at another player's card and becomes that role, immediately performing that role's night action. The Doppelgänger's win condition becomes that of the role they copied.
    \end{itemize}

    \tcbline
    
    \item \textbf{The Night Phase: Strict Action Order}
    
    All night actions happen in a fixed, silent sequence. The order is:
    Doppelgänger, Werewolves, Minion, Masons, Seer, Robber, Troublemaker, Drunk, Insomniac

    \item \textbf{The Day Phase \& Voting}
    \begin{itemize}
        \item After the Night Phase, all players discuss.
        \item The player(s) with the most votes is executed and their final role card is revealed.
    \end{itemize}
\end{enumerate}

\tcbline
\subsection*{Your Task}
For the \textbf{current statement} provided below, you must determine its strategic category and its veracity.

\end{tcolorbox}
\endgroup
\vspace{-2ex}
\captionof{figure}{The complete prompt template used to evaluate MLLMs on the MIDA task. The full list of roles and rules was provided to the models during the experiments (Main Task - Part 1).}
\label{fig:appendix_maintask_1}
\end{figure*}

\begin{figure*}[t!]
\begingroup
\scriptsize
\begin{tcolorbox}[
    colback=black!5!white, colframe=black, coltitle=white, colbacktitle=black!75!gray,
    fonttitle=\bfseries, title=Prompt for MIDA Evaluation (Main Task - Part 2),
    center title,
    arc=2mm, boxrule=0.5mm
]

\subsection*{Your Goal}
For the \textbf{current statement}, you must determine its category and veracity.

\tcbline
\subsection*{Conversation History So Far}
\texttt{{history\_str}}

\tcbline

\subsection*{Current Task}
At timestamp \texttt{{timestamp}}, \textbf{\texttt{{current\_speaker}}} says: \textbf{``\texttt{{current\_statement}}"}

Based on the rules and conversation history, please label the current statement.

\tcbline

\subsection*{Label 1: Category (Binary Classification for Each Strategy)}
For each of 6 strategy categories, indicate whether the statement contains that strategy (TRUE) or not (FALSE):
\begin{itemize}
    \item \textbf{Identity Declaration}: Directly declares identity. [TRUE/FALSE]
    \item \textbf{Accusation}: Accuses another player of being a Werewolf or lying. [TRUE/FALSE]
    \item \textbf{Defense}: Defends oneself against a question or accusation. [TRUE/FALSE]
    \item \textbf{Evidence}: Shares information obtained (or claimed) during the night. [TRUE/FALSE]
    \item \textbf{Interrogation}: Asks a question to gather information. [TRUE/FALSE]
    \item \textbf{Call for Action}: Proposes a course of action (e.g., voting). [TRUE/FALSE]
\end{itemize}
Note: If a statement is labeled as FALSE for ALL strategy categories, it is considered "No Strategy" (non-strategic).

\tcbline
\subsection*{Label 2: Veracity}
\begin{itemize}
    \item \textbf{TRUE}: The speaker knows with certainty that what they are saying aligns with the facts they currently possess.
    \item \textbf{FALSE}: The speaker knows with certainty that what they are saying contradicts the facts they currently possess, i.e., they are actively lying.
    \item \textbf{NEUTRAL}: The truthfulness of the statement cannot be determined based on the speaker's known information.
\end{itemize}

\tcbline
\subsection*{Final Output Format}
Provide your answer in the following JSON format. Do not add any text outside the JSON block.
\begin{verbatim}
{
  "identity_declaration": "TRUE or FALSE",
  "accusation": "TRUE or FALSE",
  "defense": "TRUE or FALSE",
  "evidence": "TRUE or FALSE",
  "interrogation": "TRUE or FALSE",
  "call_for_action": "TRUE or FALSE",
  "veracity": "Choose one: TRUE, FALSE, or NEUTRAL.",
  "reasoning": "Provide a brief justification for your choices 
  based on the rules and context."
}
\end{verbatim}

\end{tcolorbox}
\endgroup
\captionof{figure}{The complete prompt template used to evaluate MLLMs on the MIDA task (Main Task - Part 2).}
\label{fig:appendix_maintask_2}
\end{figure*}

\subsection{Details for MIDA Evaluation}
\label{appendix:evaluation_details}

Appendix Figure~\ref{fig:appendix_maintask_1},~\ref{fig:appendix_maintask_2} illustrate the full prompt used to evaluate MLLMs on the hierarchical MIDA task. The evaluation pipeline mirrors the structure of our annotation protocol and provides each model with the same information that human annotators rely on, ensuring a fair and reproducible comparison.

The first part of the prompt (Main Task -- Part 1) presents the complete rules of \emph{One Night Ultimate Werewolf}, including faction objectives, night action order, and role-specific abilities. As shown in the figure, these rules emphasize two principles central to MIDA: (1) players possess strictly limited private information, and (2) a player’s final role after night actions determines the truthfulness of their statements. By supplying models with this rulebook, we ensure that their judgments reflect the same deterministic game mechanics that ground our dataset’s veracity labels.

The second part of the prompt (Main Task -- Part 2) specifies the evaluation task itself. Models are provided with the full conversation history, the timestamped utterance under consideration, and are required to output two sets of predictions: (i) a six-way binary classification over persuasive strategies (Identity Declaration, Accusation, Defense, Evidence, Interrogation, Call for Action), and (ii) a final veracity label drawn from \{TRUE, FALSE, NEUTRAL\}. The output must follow a fixed JSON schema, including a brief justification grounded in the supplied rules and dialogue context. This structured format ensures that models cannot rely on generic pattern matching; instead, they must demonstrate strategy recognition and reasoning about private knowledge states.

Together, these prompts define a controlled and transparent evaluation framework. Every model receives the same rule set, the same dialogue context, and the same structured prediction requirements. This design guarantees that performance differences arise from the models’ reasoning capabilities rather than prompt variability, enabling rigorous comparison across both open- and closed-source MLLMs.

\section{Details for SoCoT and DSEM Modules}
\label{appendix:socot_dsem}

\subsection{Details for SoCoT Pipeline}
\label{appendix:socot_pipeline}

The SoCoT and DSEM modules rely on a set of structured, modality-specific prompts designed to decompose multimodal inputs into interpretable reasoning steps. Appendix Figures~\ref{fig:appendix_face_cot}–\ref{fig:high_level_reasoning_prompt} present the full templates used to guide models through this process. Together, these prompts transform raw audiovisual signals into symbolic primitives and high-level social inferences.

\noindent\textbf{Low-level Multimodal Primitives (Face, Body, Voice)}
To support Step 1 of SoCoT, we provide separate prompts for each modality:

\noindent{\textit{(1) Face Primitives.}}  
As shown in Appendix Figure~\ref{fig:appendix_face_cot}, the face-extraction prompt requires models to convert a still face crop into a structured JSON object containing a dominant emotion, a full distribution over emotion scores, localized cues (eyes, eyebrows, mouth), and a one-sentence description. This format ensures consistent grounding in concrete visual signals rather than subjective impressions.

\noindent{\textit{(2) Body Primitives.}}  
Appendix Figure~\ref{fig:appendix_body_cot} specifies the JSON schema for analyzing posture, gesture, gaze, demeanor, and holistic state. The model must describe the posture and actions, infer the overall demeanor from bodily cues, and provide supporting evidence. This yields body-level behavioral primitives compatible with SoCoT’s downstream reasoning.

\noindent{\textit{(3) Voice Primitives.}}  
Appendix Figure~\ref{fig:appendix_acoustic_cot} details the standardized mapping from objective acoustic features—intensity, pitch, and jitter—into descriptive labels such as arousal level, vocal stability, and characteristic vocal attributes. The emphasis is on producing an interpretable representation of “manner of speech” without inferring emotions beyond what the data supports.

\noindent\textbf{High-level Social Reasoning}
Appendix Figure~\ref{fig:high_level_reasoning_prompt} presents the prompt for SoCoT’s high-level reasoning stage. The model receives multimodal primitives from the Face/Body/Voice extractors and the conversation history.

Then the model is prompted to infer the speaker's likely knowledge state, possible motives, fact alignment, and potential social targets, and to synthesize these cues into a concise social inference summary. This step operationalizes the social-cognitive reasoning required for deception detection.

In practice, we combine social reasoning and decision generation into a single step for the experiment.

\subsection{Details for DSEM Module}
\label{appendix:dsem_module}

The Dynamic Social Epistemic Memory (DSEM) module maintains a persistent and incrementally updated representation of each player’s epistemic state throughout the game. Appendix Figure~\ref{fig:werewolf_knowledge_extractor} provides the full prompt used to implement the Memory-Board Extractor, which operationalizes DSEM by updating per-player knowledge strictly based on evidence from each new utterance.

\noindent\textbf{Memory Structure.}  
DSEM represents the game's evolving epistemic landscape using a set of per-player JSON files, each corresponding to an individual “Memory Board.” These boards store structured information about a player, including:
\begin{itemize}
    \item initial claims or inferred \texttt{initial\_role};  
    \item \texttt{night\_actions} performed or described;  
    \item \texttt{observed\_info} known only to that role;  
    \item \texttt{daytime\_heard} summaries;  
    \item accumulated \texttt{known\_facts} grounded in rules;  
    \item \texttt{connections} indicating relationships such as accusations or support;  
    \item \texttt{reaction\_notes} and \texttt{temporal\_patterns};  
    \item a short \texttt{history} of events related to the player.
\end{itemize}

Each game session (\texttt{game\_id}) is stored in a dedicated directory, with one JSON file per participant. This ensures that DSEM remains modular, interpretable, and easily inspectable.

\noindent\textbf{Turn-by-turn incremental updates.}  
Whenever a new statement is spoken, DSEM triggers an update cycle. The Memory-Board Extractor prompt takes as input:
\begin{itemize}
    \item the current utterance and its timestamp,  
    \item the speaker identity,  
    \item the prior conversation history, and  
    \item the relevant snippet of the current knowledge board.
\end{itemize}

The model is instructed to produce a strict JSON structure containing an \texttt{"updates"} list. Each update may add or remove facts, append night-action claims, register observed information, or create new interpersonal connections grounded in evidence. Importantly, the prompt enforces \emph{conservative updating}: fields must be left \texttt{Unknown} when the evidence is insufficient, ensuring that DSEM does not hallucinate global inferences or introduce unsupported beliefs.

These updates are merged back into the persistent per-player JSON files. This incremental mechanism ensures that the epistemic state remains synchronized with the flow of dialogue and evolves only when justified by new statements.

\noindent\textbf{Using DSEM during evaluation.}  
For the main MIDA reasoning task, the system retrieves the most recent snapshot of a player’s Memory Board to construct a concise ``Snippet.” This snippet provides the model with distilled epistemic information, such as recent known facts or relevant connections, while allowing configurable truncation.  

The snippet is then injected directly into the evaluation prompt (Appendix A.2), where it serves as structured context alongside the conversation history and multimodal cues. This allows the model to ground its strategic and deception judgments not only in the transcript but also in a dynamically maintained representation of each participant’s evolving knowledge and social relationships.

\begin{figure*}[t!]
\begin{tcolorbox}[
    colback=black!5!white, colframe=black, coltitle=white, colbacktitle=black!75!gray,
    fonttitle=\bfseries, title=Prompt for Face Primitives Extraction,
    center title,
    arc=2mm, boxrule=0.5mm
]

Please perform a detailed analysis of the facial expression in the provided image and return your analysis in a strict JSON format. Do not include any text outside of the JSON object.

\tcbline

\textbf{Required JSON Format:}
The JSON object must have the following keys:
\begin{itemize}
    \itemsep0em
    \item \texttt{"dominant\_emotion"}: A string representing the primary emotion with the highest score from the \texttt{emotion\_scores} object below.
    \item \texttt{"emotion\_scores"}: An object where you must assign a score to \textbf{each} of the predefined emotion categories listed below. The score should be a float from 0.0 to 1.0, where 0.0 means the emotion is completely absent, and 1.0 means the emotion is extremely prominent and intensely expressed. The emotion categories are:
    \begin{itemize}
        \itemsep0em
        \item \texttt{"happiness"}
        \item \texttt{"sadness"}
        \item \texttt{"anger"}
        \item \texttt{"surprise"}
        \item \texttt{"fear"}
        \item \texttt{"disgust"}
        \item \texttt{"neutral"}
        \item \texttt{"contempt"}
        \item \texttt{"contemplation"}
    \end{itemize}
    \item \texttt{"facial\_cues"}: An object detailing the specific facial features supporting your scoring, with keys like \texttt{"eyes"}, \texttt{"eyebrows"}, and \texttt{"mouth"}.
    \item \texttt{"description"}: A brief, one-sentence summary of the overall emotional expression.
\end{itemize}

\tcbline

\textbf{Task:} Now, analyze the image and provide the complete JSON output.

\end{tcolorbox}
\captionof{figure}{The complete prompt used in the SoCoT pipeline specifically on facial expressions.}
\label{fig:appendix_face_cot}
\end{figure*}

\begin{figure*}[t!]
\begin{tcolorbox}[
    colback=black!5!white, colframe=black, coltitle=white, colbacktitle=black!75!gray,
    fonttitle=\bfseries, title=Prompt for Body Primitives Extraction,
    center title,
    arc=2mm, boxrule=0.5mm
]

Please perform a detailed analysis of the person's overall body posture, gesture, demeanor, and facial expression in the provided image, and return your analysis strictly in JSON format. Do not include any text outside the JSON object.

\tcbline

\textbf{Required JSON Format:}
The JSON object must contain the following fields:
\begin{itemize}
    \itemsep0em
    \item \texttt{"dominant\_posture"}: A string describing the person's main body posture (e.g., "sitting", "standing", "leaning forward", "leaning back", etc.).
    \item \texttt{"body\_action"}: A string describing the person's current body gesture or action (e.g., "pointing at someone", "arms crossed", "emphatic hand gesture", "shrugging", "arms folded", etc.).
    \item \texttt{"facial\_expression"}: A string briefly describing the facial expression (e.g., "smiling", "frowning", "surprised", "neutral", etc.).
    \item \texttt{"gaze\_direction"}: A string describing the gaze direction (e.g., "looking to the left at someone", "looking directly at the camera", "looking down", "looking around", etc.).
    \item \texttt{"emotion"}: A string summarizing the overall emotion based on body and facial cues (e.g., "confident", "nervous", "relaxed", "confused", "angry", "happy", etc.).
    \item \texttt{"description"}: A one-sentence summary of the person's overall state and possible psychological activity.
    \item \texttt{"supporting\_evidence"}: A brief explanation of the basis for your judgments above, indicating how you inferred from body, posture, and facial features.
\end{itemize}

\tcbline

\textbf{Task:} Now, analyze the image and output the complete JSON object.

\end{tcolorbox}
\captionof{figure}{The complete prompt used in the SoCoT pipeline on holistic body language, including posture, gestures, and expression.}
\label{fig:appendix_body_cot}
\end{figure*}

\begin{figure*}[t!]
\begin{tcolorbox}[
    colback=black!5!white, colframe=black, coltitle=white, colbacktitle=black!75!gray,
    fonttitle=\bfseries, title=Prompt for Voice Primitives Extraction,
    center title,
    arc=2mm, boxrule=0.5mm
]

You are a professional Acoustic and Speech Analyst. Your task is to convert objective acoustic feature data into a set of standardized attribute labels describing the ``manner of delivery'' of the voice. Please avoid making strong or specific emotional judgments, focusing strictly on objective description.

\tcbline

\textbf{Background Knowledge and Reference Range:}
\begin{itemize}
    \itemsep0em
    \item \textbf{Intensity}: Reflects loudness. Normal speech is approximately 0.01. High values indicate loudness.
    \item \textbf{Pitch Mean (in Hz)}: Reflects fundamental frequency. Males approx. 85--180Hz, females approx. 165--255Hz.
    \item \textbf{Jitter}: Reflects voice stability. $<0.01$ indicates stability. $>0.02$ indicates instability or tension.
\end{itemize}

\textbf{Acoustic Data for Analysis:}
\begin{itemize}
    \itemsep0em
    \item Intensity: \{intensity:.6f\}
    \item Pitch Mean: \{pitch\_mean:.2f\} Hz
    \item Jitter: \{jitter:.6f\}
\end{itemize}

\tcbline

\textbf{Task Instructions:}
Analyze the data above and strictly generate a JSON object. Your goal is to provide a set of ``labels,'' not a ``conclusion.'' The JSON object must contain the following fields:

\begin{itemize}
    \itemsep0em
    \item \texttt{"arousal"}: String. Physiological arousal level. Must be one of ["Calm", "Normal", "Elevated"].
    \item \texttt{"stability"}: String. Voice stability. Must be one of ["Stable", "Slightly Unstable", "Unstable"].
    \item \texttt{"vocal\_attributes"}: Array of strings. Select descriptive labels from the preset list below: \\
    \textit{("Loud", "Quiet", "High-Pitched", "Low-Pitched", "Monotonous", "Expressive", "Shaky / Tense", "Clear / Steady", "Energetic", "Subdued")}
    \item \texttt{"summary"}: String. A purely descriptive summary, avoiding specific emotional vocabulary.
\end{itemize}

\tcbline

\textbf{Example Output:}
\begin{itemize}
    \itemsep0em
    \item Input Data: \\ Intensity: 0.25, Pitch Mean: 310.5 Hz, Jitter: 0.025
    \item Output JSON: \\ \texttt{\{"arousal": "Elevated", "stability": "Unstable", "vocal\_attributes": ["Loud", "High-Pitched", "Shaky / Tense", "Energetic"], "summary": "..."\}}
\end{itemize}

\end{tcolorbox}
\captionof{figure}{The prompt used to guide the model in converting voice features (Intensity, Pitch, Jitter) into standardized descriptive labels.}
\label{fig:appendix_acoustic_cot}
\end{figure*}

\begin{figure*}[t!]
\begin{tcolorbox}[
    colback=black!5!white, colframe=black, coltitle=white, colbacktitle=black!75!gray,
    fonttitle=\bfseries, title=Prompt for High-Level Social Reasoning,
    center title,
    arc=2mm, boxrule=0.5mm
]

You will produce \textbf{High-Level Social Reasoning} for ``One Night Ultimate Werewolf''. Use the conversation so far and the multi-modal analyses to infer the social meaning.

\tcbline

\textbf{Inputs/Context:}
\begin{itemize}
    \itemsep0em
    \item \textbf{Conversation History}: \texttt{\{history\_str\}}
    \item \textbf{Current Speaker}: \texttt{\{speaker\}}
    \item \textbf{Current Statement}: \texttt{"\{statement\}"}
\end{itemize}

\textbf{Analyses JSON Context:}

The model receives a combined JSON object named \texttt{"context"} integrating low-level inputs:
\begin{itemize}
    \itemsep0em
    \item \texttt{body} / \texttt{face} / \texttt{voice} (Multimodal Analysis Results)
\end{itemize}
\tcbline

\textbf{Required JSON Output Structure:}
Return \textbf{ONLY} a JSON object with the following fields:

\begin{itemize}
    \itemsep0em
    \item \texttt{"knowledge\_state"}: What the Speaker (A) likely knows now (cautious inference).
    \item \texttt{"possible\_motives"}: Array of concise hypotheses (e.g., ``mislead B'', ``pressure C to reveal'').
    \item \texttt{"statement\_fact\_alignment"}: One of [``Aligned'', ``Contradictory'', ``Unknown''], plus short reason.
    \item \texttt{"potential\_targets"}: Who is being influenced and how (if inferable), else [].
    \item \texttt{"risk\_assessment"}: Brief risks for the Speaker (A) given the claim.
    \item \texttt{"social\_inference\_summary"}: One-sentence synthesis.
\end{itemize}

\end{tcolorbox}
\captionof{figure}{The prompt used for the model social reasoning, integrating multimodal analyses to infer social motives, knowledge, and alignment within the game context.}
\label{fig:high_level_reasoning_prompt}
\end{figure*}

\begin{figure*}[t!]
\begin{tcolorbox}[
    colback=black!5!white, colframe=black, coltitle=white, colbacktitle=black!75!gray,
    fonttitle=\bfseries, title=Prompt for DSEM Memory-Board Extraction,
    center title,
    arc=2mm, boxrule=0.5mm
]

You are a \textbf{Memory-Board Extractor} for the game ``One Night Ultimate Werewolf''. Your job is to update per-player knowledge states \textbf{ONLY} based on evidence from the current statement and minimal inference grounded in rules and history. Avoid global speculation.

\tcbline

\textbf{Inputs/Context:}
\begin{itemize}
    \itemsep0em
    \item \textbf{Timestamp}: \texttt{\{timestamp\}}
    \item \textbf{Current Speaker}: \texttt{\{speaker\}}
    \item \textbf{Current Statement}: \texttt{"\{statement\}"}
    \item \textbf{Conversation History} (previous turns): \texttt{\{history\_str\}}
    \item \textbf{Current Memory Board Snippet} (for reference): \\ \texttt{\{json.dumps(board\_snippet, ensure\_ascii=False)\}}
\end{itemize}

\tcbline

\textbf{Required JSON Output Structure:}
Return a \textbf{STRICT JSON} object with the format containing an \texttt{"updates"} array of player update objects.

\begin{itemize}
    \itemsep0em
    \item \texttt{"updates"}: [
    \begin{itemize}
        \itemsep0em
        \item \texttt{"player"}: Name of player to update.
        \item \texttt{"initial\_role"}: ``Seer$|$Werewolf$|$Villager$|\dots|$Unknown$|$null''
        \item \texttt{"initial\_role\_basis"}: Optional short basis.
        \item \texttt{"night\_actions"}: List of inferred night actions (e.g., ``looked at card X'').
        \item \texttt{"observed\_info"}: List of observations (e.g., ``saw player B is Villager'').
        \item \texttt{"daytime\_heard"}: List of statements heard.
        \item \texttt{"known\_facts\_add"}: List of cautious, evidence-grounded facts to add.
        \item \texttt{"known\_facts\_remove"}: List of facts to remove.
        \item \texttt{"connections"}: Array of player relationships (\texttt{from}, \texttt{to}, \texttt{relation}, \texttt{basis}).
        \item \texttt{"reaction\_notes"}: Array of notes on player reactions (\texttt{player}, \texttt{note}, \texttt{confidence}).
        \item \texttt{"temporal\_patterns"}: List of patterns inferred across time.
        \item \texttt{"history\_event"}: Optional short event summary for this timestamp.
    \end{itemize}
    ]
\end{itemize}

\tcbline

\textbf{Rules:}
\begin{itemize}
    \itemsep0em
    \item Only include fields you can justify from the statement/history.
    \item Prefer adding to \texttt{known\_facts} via cautious, evidence-grounded assertions.
    \item If unsure, use \texttt{Unknown/null} and skip fields.
\end{itemize}

\end{tcolorbox}
\captionof{figure}{The complete prompt used to instruct a model to act as a Memory-Board Extractor, updating player states based on turn-by-turn evidence in the game ``One Night Ultimate Werewolf.''}
\label{fig:werewolf_knowledge_extractor}
\end{figure*}

\end{document}